\documentclass[3p]{elsarticle}
\makeatletter
\def\ps@pprintTitle{%
 \let\@oddhead\@empty
 \let\@evenhead\@empty
 \def\@oddfoot{\centerline{\thepage}}%
 \let\@evenfoot\@oddfoot}
\makeatother

\usepackage{lineno,hyperref}
\modulolinenumbers[0]

\journal{Engineering Applications of Artificial Intelligence}

\biboptions{authoryear}

\usepackage{xcolor}
\usepackage{graphicx}
\usepackage{subcaption} 
\usepackage{amsmath}
\usepackage{amsfonts}
\usepackage{amssymb}
\usepackage{algorithm}
\usepackage{booktabs}
\usepackage{multirow}
\usepackage{threeparttable}
\usepackage{algorithmicx}
\usepackage{algpseudocode}
\newsavebox\CBox
\def\textBF#1{\sbox\CBox{#1}\resizebox{\wd\CBox}{\ht\CBox}{\textbf{#1}}}

\begin{document}

\begin{frontmatter}

\title{Learning Variable Ordering Heuristics for Solving Constraint Satisfaction Problems\tnoteref{mytitlenote}}
\tnotetext[mytitlenote]{Accepted by Engineering Applications of Artificial Intelligence. DOI: https://doi.org/10.1016/j.engappai.2021.104603}


\author[label1]{Wen Song}
\ead{wensong@email.sdu.edu.cn}

\author[label2]{Zhiguang Cao\corref{mycorrespondingauthor}}
\cortext[mycorrespondingauthor]{Corresponding author}
\ead{zhiguangcao@outlook.com}

\author[label3]{Jie Zhang}
\ead{zhangj@ntu.edu.sg}

\author[label2]{Chi Xu}
\ead{cxu@simtech.a-star.edu.sg}

\author[label4]{Andrew Lim}
\ead{i@limandrew.org}

\address[label1]{Institute of Marine Science and Technology, Shandong University, China}
\address[label2]{Singapore Institute of Manufacturing Technology (SIMTech), Singapore}
\address[label3]{School of Computer Science and Engineering, Nanyang Technological University, Singapore}
\address[label4]{School of Computing and Artificial Intelligence, Southwest Jiaotong University, China}

\begin{abstract}
Backtracking search algorithms are often used to solve the Constraint Satisfaction Problem (CSP), which is widely applied in various domains such as automated planning and scheduling. The efficiency of backtracking search depends greatly on the variable ordering heuristics. Currently, the most commonly used heuristics are hand-crafted based on expert knowledge. In this paper, we propose a deep reinforcement learning based approach to automatically discover new variable ordering heuristics that are better adapted for a given class of CSP instances, without the need of relying on hand-crafted features and heuristics. We show that directly optimizing the search tree size is not convenient for learning, and propose to optimize the expected cost of reaching a leaf node in the search tree. To capture the complex relations among the variables and constraints, we design a representation scheme based on Graph Neural Network that can process CSP instances with different sizes and constraint arities. Experimental results on random CSP instances show that on small and medium sized instances, the learned policies outperform classical hand-crafted heuristics with smaller search tree (up to 10.36\% reduction). Moreover, without further training, our policies directly generalize to instances of larger sizes and much harder to solve than those in training, with even larger reduction in the search tree size (up to 18.74\%).
\end{abstract}


\end{frontmatter}


\section{Introduction}
\label{sec:introduction}
Combinatorial problems widely exist in many domains \citep{petit2019enriching}. 
As one of the most commonly studied combinatorial problems in computer science and artificial intelligence, Constraint Satisfaction Problem (CSP) provides a general framework for modeling and solving combinatorial problems. Solving CSP plays a central role in many research areas. A typical and successful application of CSP is automated planning and scheduling \citep{salido2008introduction}, which has numerous real-world applications, ranging from industry 4.0 \citep{legat2017configurable,song2019sampling} to robotics \citep{kasprzak2014hierarchical,behrens2019constraint}. A CSP instance involves a set of variables and constraints. To solve it, one needs to find a value assignment (i.e. solution) for all the variables such that all the  constraints are satisfied, or prove such assignment does not exist. Despite its ubiquitous applications, unfortunately, CSP is well known to be NP-complete in general \citep{mackworth1993complexity}. To solve CSP efficiently, \emph{backtracking search} algorithms are often employed, which are exact algorithms with the guarantee that a solution will be found if one exists. Though the worst-case complexity is still exponential, with the help of constraint propagation, backtracking search algorithms often perform reasonably well in practice \citep{rossi2006handbook}.

In general, a backtracking search algorithm performs depth-first traverse of a search tree, and tries to find a solution by iteratively selecting a variable and applying certain branching strategy. The decision of which variable to select next is referred to as \emph{variable ordering}. It is well acknowledged that the choice of variable ordering has a critical impact on the efficiency of backtracking search algorithms \citep{gent1996empirical}. However, finding the optimal ordering, i.e. the one results in a minimal-sized search tree, is at least as hard as solving the CSP \citep{liberatore2000complexity,rossi2006handbook}. Therefore, current practice mainly relies on hand-crafted variable ordering heuristics obtained from the experience of human experts, such as \textsf{MinDom} \citep{haralick1980increasing}, \textsf{Dom/Ddeg} \citep{bessiere1996mac}, and impact-based heuristic \citep{refalo2004impact}. Though they are easy to use and widely adopted, they do not have any formal guarantees on the optimality. In addition, they are designed for solving any CSP instance without considering the problem-specific structures, which can be exploited to achieve much better efficiency. However, incorporating these additional features requires substantial experience and deep domain knowledge, which are hard to obtain in reality \citep{rossi2006handbook}. 

Recently, Deep Neural Networks (DNNs) have been shown to be promising in learning algorithms for solving NP-hard problems, such as routing, graph problems, Propositional Satisfiability Problem (SAT), and so on \citep{bengio2020machine}. The effectiveness comes from the fact that given a class of problem instances (e.g. drawn from a distribution), DNN can be trained to discover useful patterns that may not be known or hard to be specified by human experts, through supervised or reinforcement learning (RL) \citep{ma2020cost,altan2021new}. Based on the above success, in this paper, we ask the following question: \emph{can we use DNN to discover better variable ordering heuristics for a class of CSP?} This is not a trivial task, due to the following challenges. Firstly, given the exponential (worst-case) complexity of CSP, it is not practical to obtain large amount of labeled training data (e.g. optimal search paths), therefore it is hard to apply supervised learning methods. Secondly, CSP instances have different sizes in terms of number of variables and constraints, and the constraint arities (i.e. number of variables involved) are also different. It is crucial to design a deep representation scheme that can effectively process CSP instances of any size and constraint arity.

In the literature, several works have tried to leverage machine learning techniques to learn variable ordering heuristics for solving satisfaction problems, such as SAT \citep{lagoudakis2001learning}, Quantified Boolean Formulas (QBF) \citep{samulowitz2007learning}, and CSP \citep{epstein2007learning,xu2009learning}. However, as will be detailed in Section \ref{sec:relatedWork}, these methods cannot directly generate ordering policies from solving status. They require a set of predefined heuristics from which the algorithm learns to choose (or combine), and are all based on conventional learning methods and hand-crafted features. More recently, independent to our work, DNN is used with RL to learn value ordering heuristics for CSP \citep{cappart2021combining,chalumeau2021seapearl}. However, as mentioned in \citep{chalumeau2021seapearl}, learning variable ordering heuristics with DNN raises additional challenges to the learning mechanism.

In this paper, we design a deep reinforcement learning (DRL) agent which tries to make the optimal variable ordering decisions at each decision point to minimize the search tree size. More specifically, variable ordering in backtracking search is modeled as a Markov Decision Process (MDP), where the optimal policy is to select at each decision point the variable with the minimum expected number of search nodes. The DRL agent can optimize its policy by learning from its own experiences of solving CSP instances drawn from a distribution, without the need of supervision. However, as will be shown later, such direct formulation could cause inefficiency and inconvenience to the learning mechanism, since learning must be delayed until backtracking from a search node. To resolve this issue, we consider the search paths originated from a node as separate trajectories, and opt to minimize the expected number of remaining nodes to reach a leaf node. We represent the internal states of the search process based on Graph Neural Network (GNN) \citep{xu2018powerful}, which can process CSP instances of any size and constraint arity, and effectively capture the relationship between the variables and constraints. We use Double Deep Q-Network (DDQN) \citep{van2016deep} to train the GNN based RL agent. Experimental results on random CSP instances generated by the well-known model RB \citep{xu2007random} show that the RL agent can discover policies that are better than the traditional hand-crafted variable ordering heuristics, in terms of minimizing the search tree size. More importantly, the learned policy can effectively \emph{generalize} to larger instances that have never been seen during training.

To summarize, the contributions of this paper are as follows:
\begin{itemize}
    \item We propose a novel DRL method to learn variable ordering heuristics for solving a class of CSP. Different from previous works, our method learns in an end-to-end fashion, meaning that through self-training, it can automatically generate high-quality ordering heuristics from raw state input without the need of hand-crafted features and predefined heuristics.
    \item We propose to minimize the expected search effort of reaching the leaf nodes, instead of directly minimizing the search tree size. In this way, RL training could be more efficient since temporal difference learning is effectively enabled. Specifically, the agent can learn at each search step, instead of waiting until backtracking.
    \item We propose a GNN based scheme to represent the internal search states, based on which a deep Q network is designed to output the Q-value of each candidate variable end-to-end from raw state features of variables and constraints. Such representation is agnostic to the instance size, hence enables generalizing to large unseen instances.
\end{itemize}

The rest of this paper is organized as follows. We first summarize related works in Section \ref{sec:relatedWork}, followed by preliminaries about CSP in Section \ref{sec:pre}. Then in Section \ref{sec:method}, we present our deep RL method in detail. Section \ref{sec:experiments} provides the experimental results and analysis. Section \ref{sec:discussion} further discusses the reasons of the good performance of our method, as well as the limitations. Finally, Section \ref{sec:conclusions} concludes the paper.

\section{Related Work} \label{sec:relatedWork}
Recently, there has been an increasing attention on using deep (reinforcement) learning to tackle hard combinatorial (optimization or satisfaction) problems. Quite a few works in this direction focus on solving specific types of problems, including routing  \citep{kool2018attention,wu2021learning,xin2021step,xin2021multi}, graph problems \citep{khalil2017learning,li2018combinatorial}, 
and scheduling \citep{zhang2020learning,mao2019learning}. Instead of solving specific problems, we focus on CSP which is a general representation of combinatorial problems.

In the literature, a number of methods try to tackle satisfaction problems such as CSP and SAT in an end-to-end fashion, meaning that training DNN to directly output a solution for a given instance. \cite{xu2018towards} represent binary CSP as a matrix and train a Convolutional Neural Network (CNN) to predict its satisfiability, but cannot give the solution for satisfiable instances. In addition, the matrix representation scheme cannot scale to arbitrary problem size. \cite{galassi2018model} train a DNN that can construct a feasible solution of a CSP instance by extending a partial assignment, however the representation scheme based on one-hot encoding of assignment also suffers from the issue of being restricted to a pre-determined problem size. \cite{selsam2018learning} train a satisfiability classifier for SAT, which can be considered as a special case of CSP. The underlying architecture is based on GNN instead of CNN, therefore can process instances with different sizes. The authors also provide an unsupervised procedure to decode a satisfying assignment. \cite{amizadeh2018learning} propose a differentiable architecture to train a GNN that directly aiming at solving the Circuit-SAT problem instead of only predicting its satisfiability.

Despite their simplicity and effectiveness, as pointed out by Bengio et al. in a recent survey \citep{bengio2020machine}, end-to-end methods suffer from two major limitations: 1) feasibility is weak since it is hard for them to handle advanced types of constraints, and 2) no guarantee on the solution quality (in terms of optimality and feasibility for optimization and satisfaction problems, respectively). A more promising way is to apply machine learning within the framework of exact algorithms, such that the feasibility and solution quality can be guaranteed \citep{bengio2020machine}. A typical exact framework is the branch-and-bound algorithm for solving Mixed Integer Linear Programs (MILPs). \cite{he2014learning} use imitation learning to learn a control policy for selecting branches in the branch-and-bound process. \cite{khalil2016learning} achieves similar purpose by solving a learning-to-rank task to mimic the behaviors of strong branching. \cite{khalil2017learningRun} also develop a machine learning model to decide whether the primal heuristics should be run for a given branch-and-bound node. These methods are based on linear models with static and dynamic features describing the current branch-and-bound status. More recently, \cite{gasse2019exact} use imitation learning to mimic strong branching, where the underlying states are represented using GNN. Similarly, a GNN based network is designed in \citep{ding2020accelerating}, which is trained in a supervised way to predict values of binary variables in MILP. Though sharing similar GNN structure, our work differs from \citep{gasse2019exact,ding2020accelerating} in mainly two aspects. First, our method does not require labels that are costly to obtain but necessary for imitation or supervised learning. Second, as will be shown latter, we only uses 4 simple raw features, while \citep{gasse2019exact,ding2020accelerating} rely on 19 and 22 complex MILP features, respectively.

Another exact framework is the backtracking search algorithms for solving satisfaction problems.  \cite{balafrej2015multi} use bandit model to learn a policy that can adaptively select the right constraint propagation levels at each node of a CSP search tree. More close to our work, several methods use traditional machine learning to choose the branching heuristics for solving CSP and some special cases. \cite{lagoudakis2001learning} use RL to learn the branching rule selection policy for the \#DPLL algorithm for solving SAT, which requires finding all solutions for a satisfiable instance. However, as will be discussed in Section \ref{sec:method}, this RL formulation is not directly applicable for learning in our case. \cite{samulowitz2007learning} study the heuristic selection task for solving Quantified Boolean Formulas (QBF), a generalization of SAT, through supervised learning. In terms of CSP, \cite{epstein2007learning} opt to learn a linearly weighted profile of multiple ordering heuristics to select the next variable and value for each search node. Though their training mechanism is self-supervised by using the solver's own solving experiences, it is not formulated as a RL task and the weight of each heuristic is learned simply based on the frequency it supports correct or oppose incorrect decisions. \cite{xu2009learning} propose a RL formulation for variable ordering heuristic selection, but only provide preliminary results. 

Though sharing similar goals, our approach significantly differs from \citep{lagoudakis2001learning,samulowitz2007learning,epstein2007learning,xu2009learning} in several ways. Firstly, we propose a RL formulation that are suitable for temporal difference learning during backtracking search, instead of wait until solving is complete. Secondly, in our approach, the learned policy directly picks the next variable based on its own estimates of the environment, without the need of consulting a set of predefined heuristics. Finally, our approach can leverage the approximation and expressive power of DNN. Our GNN based representation scheme provides an effective way to capture the complex relations among variables and constraints of CSP. More importantly, it can effectively process instances of arbitrary sizes and constraint arities which is not viable for the existing deep representations of CSP in \citep{xu2018towards,galassi2018model}.

Concurrent and independent to our work, \cite{cappart2021combining} formulate the CSP search process as dynamic programming, and employ deep RL to learn value ordering heuristics, which choose a value for the selected variable to initiate. This work is further extended in \citep{chalumeau2021seapearl}, which fully embeds deep reinforcement learning within the CSP solver. Different from these two works, we focus on learning variable ordering policies which is critical to the search performance and could possess unique requirements on the learning mechanism. In the literature, research on learning variable ordering heuristics based on deep RL is rather sparse \citep{popescu2021overview}.

\section{Preliminaries} \label{sec:pre}
A Constraint Satisfaction Problem (CSP) can be formally defined on a \emph{constraint network} \citep{balafrej2015multi}, which is a triple $P=<\mathcal{X},\mathcal{D},\mathcal{C}>$, where $\mathcal{X}=\{x_1,...,x_n\}$ is a set of $n$ variables indexed by $i$, $\mathcal{D}=\{d(x_1),...,d(x_n)\}$ is the domain of each $x_i$, and $\mathcal{C}=\{c_1,...,c_e\}$ is a set of $e$ constraints indexed by $j$. A domain $d(x_i)$ is a finite set of values that can be assigned to $x_i$. A constraint $c_j$ is a pair $c_j=<scp(c_j),rel(c_j)>$, where $scp(c_j)\subseteq \mathcal{X}$ is the scope of $c_j$ specifying the variables involved in $c_j$, and $rel(c_j)$ is the relation containing all the allowed value combinations (tuples) of variables in $scp(c_j)$. The cardinality of $scp(c_j)$, i.e. the number of variables involved in $c_j$, is called the arity of the constraint. In this paper, we assume $\mathcal{C}$ contains only table constraints, i.e. all the allowed tuples for a constraint are explicitly listed as a table. This is somewhat limited, but table constraints are one of the most fundamental and useful constraint types since they can theoretically represent any constraint of other type \citep{demeulenaere2016compact}. A solution to the constraint network is an assignment of all the variables such that all the constraints are satisfied. Solving a CSP is to find one solution of the constraint network\footnote{Generally, one may require to find more than one, or even all solutions, if the CSP instance is satisfiable. While we assume finding one is enough, our approach can be applied when more solutions are required.}, or prove no solution exists, i.e. the CSP is unsatisfiable. 

The backtracking search process can be considered as performing a depth-first traverse of the \emph{search tree}, which is dynamically constructed during the search process. At each node, the algorithm selects a variable from those have not been assigned a value yet (i.e. unbounded) according to some variable ordering heuristic, and then, based on certain branching strategy, generates multiple child nodes by posting a set of mutually exclusive and exhaustive branching constraints and performing \emph{constraint propagation} (CP). Essentially, CP is used to remove some values that are not consistent with the current branching decisions, which can significantly reduce the search space and is the key to achieve high computational efficiency. Hence, each search node corresponds to a \emph{subinstance} of the original CSP instance being solved, with the same constraints (ignoring branching constraints) and smaller domains. If the domain of some variable is empty after constraint propagation, then the corresponding node is a dead-end since the current branching decisions cannot lead to any feasible solution, and the algorithm backtracks. Search terminates when a solution is found, or the search tree has been completely traversed, meaning that the instance is unsatisfiable. Therefore, the leaf nodes of a search tree include dead-ends and the one with the feasible solution, if one exists.

\begin{figure*}
	\centering
	\begin{subfigure}[b]{0.32\textwidth}
		\centering
		\includegraphics[scale=0.04]{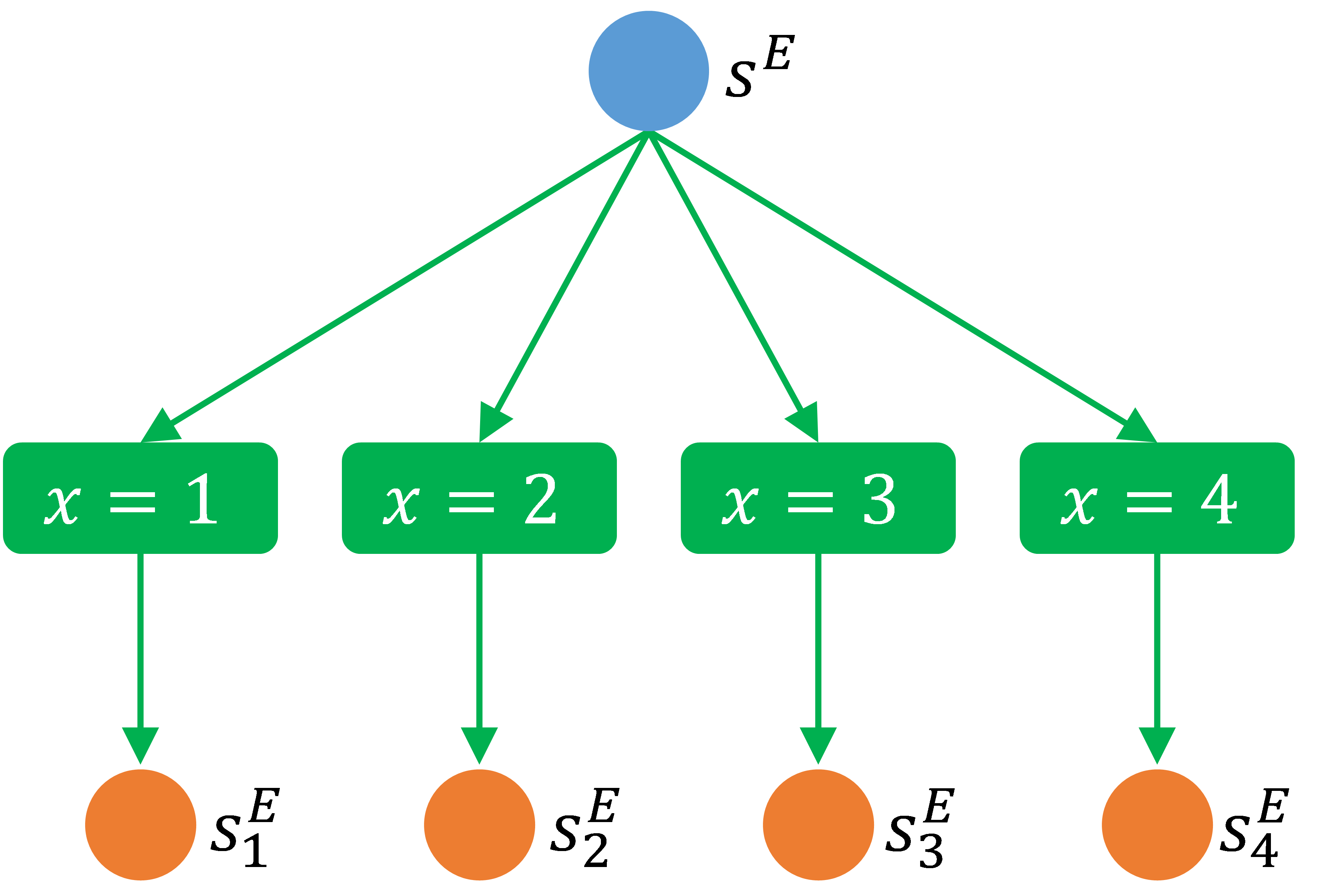}
		\caption{Enumeration}
		\label{fig:enum}
	\end{subfigure}
	\begin{subfigure}[b]{0.32\textwidth}
		\centering
		\includegraphics[scale=0.04]{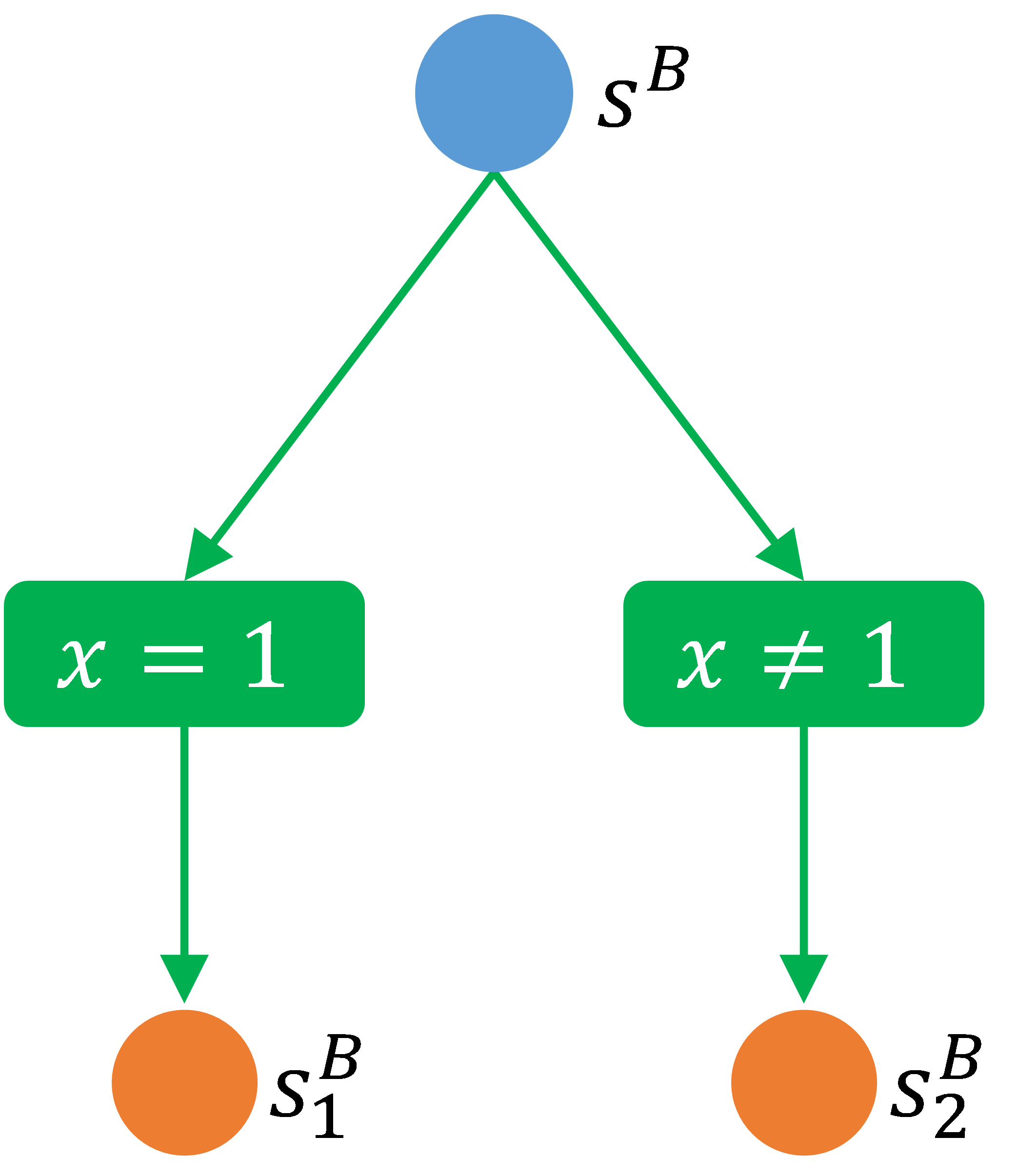}
		\caption{Binary}
		\label{fig:binary}
	\end{subfigure}
	\begin{subfigure}[b]{0.32\textwidth}
		\centering
		\includegraphics[scale=0.04]{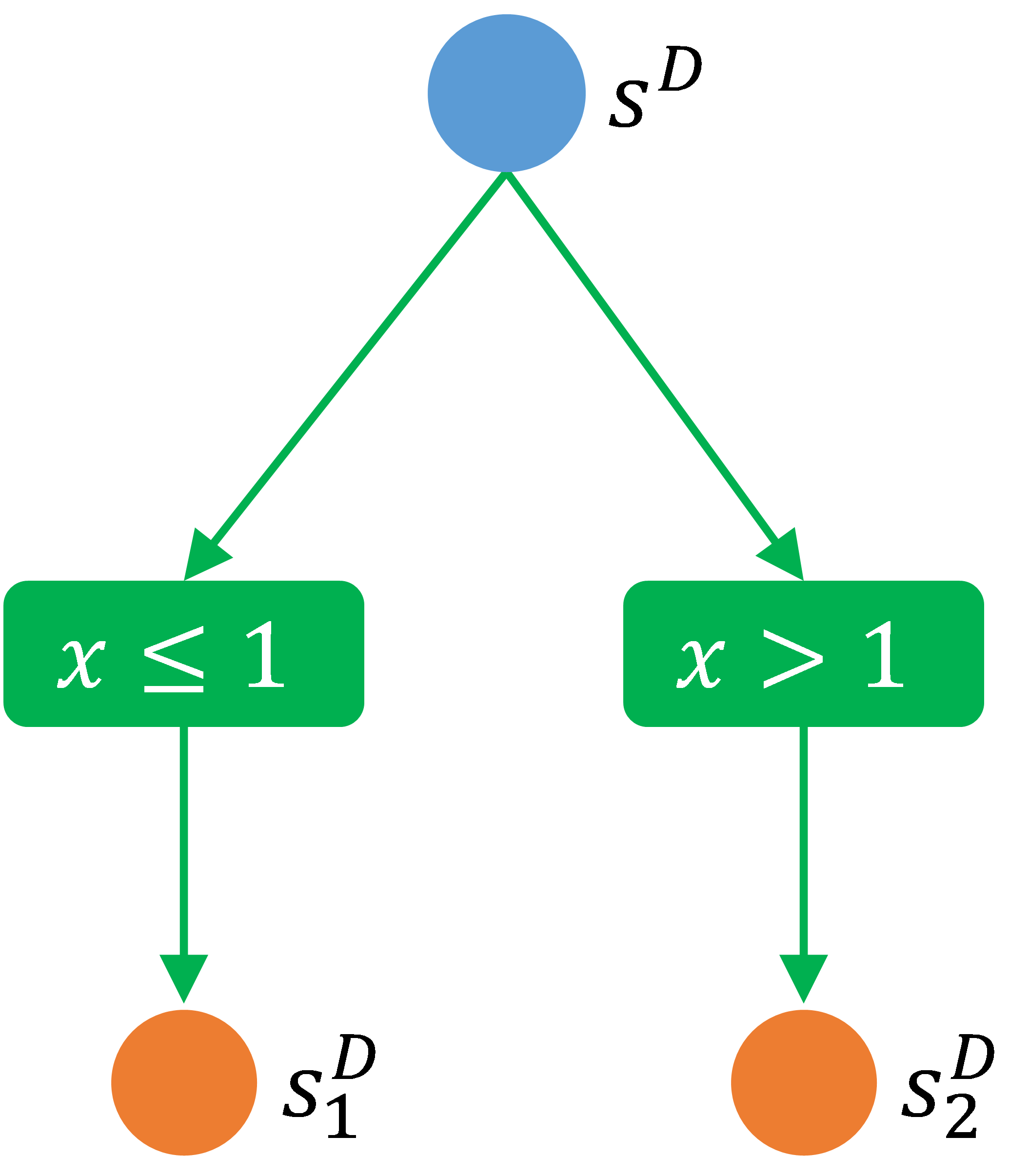}
		\caption{Domain splitting}
		\label{fig:dom_split}
	\end{subfigure}
	\caption{An illustration of three branching strategies. The blue and orange circles are parent and child nodes, respectively, and the green boxes are branching constraints. In this example, a variable $x$ with domain $d(x)=\{1,2,3,4\}$ is selected. The four values are ordered ascendingly for branching.}
	\label{fig:branching_strategy}
\end{figure*}

For backtracking search, one of the most commonly used branching strategies is enumeration, or $d$-way branching, where each child node corresponds to instantiating the selected variable $x$ with a value $l$ in its domain (i.e. posting $x=l$). The selection of $l$ is based on certain value ordering heuristic. Unlike enumeration, the binary branching strategy, or $2$-way branching, generates two children for a search node, where a value $l$ of the selected variable is applied (i.e. posting $x=l$) on the left branch and refuted (i.e. posting $x\neq l$) on the right branch. Another popular alternative is domain splitting, which posts two branching constraints based on the selected variable $x$ and value $l$, e.g. $x\leq l$ and $x>l$, to produce two children. A simple illustration of these branching strategies is shown in Figure \ref{fig:branching_strategy}. Our approach is not limited to a particular strategy, since it assumes the general search tree structure. Note that a search node is created only when it is \emph{visited} by the algorithm. For example, suppose the underlying CSP instance in Figure \ref{fig:enum} is satisfiable. When reaching the parent node $s^E$, child $s^E_1$ is visited first. If the algorithm finds out $s^E_1$ is infeasible, it backtracks to the parent and then creates $s^E_2$. If a solution is found under $s^E_2$, the algorithm terminates therefore $s^E_3$ and $s^E_4$ will not be visited.

\section{Method} \label{sec:method}
In this section, we formally describe our proposed approach. We first formulate the variable ordering heuristic discovery as a reinforcement learning task. Then, we present our GNN based state representation scheme. Finally, we describe the training algorithm in detail.

\subsection{The Reinforcement Learning Formulation}
Our goal in this paper is to train a reinforcement learning (RL) agent to perform variable ordering with the objective of minimizing the search tree size, defined as the total number of search nodes. To formulate the RL task, we first need to define the underlying Markov Decision Process (MDP), where the agent is responsible for making the variable ordering decisions, and the solver is considered as the environment. Here we define a state $s$ as the instance (for the root node) or subinstance associated with a search node. The states for the leaf nodes are defined as terminal states. Given state $s$, an action $a$ is to select an unbounded variable for branching, hence we define the action set as $A(s) = \{x_i\in \mathcal{X}| \left|d(x_i)\right|> 1 \}$. Given a simple transition $(s,a,s')$, we define the cost $r(s,a,s')=1$, meaning that one more search node is visited.

However, the actual state transition in backtracking search is not the simple ones. Since a search node could have multiple child nodes, state transitions are not one-to-one as in typical MDP, but one-to-many. Consider the example in Figure \ref{fig:enum}, the state on the parent $s^E$ will transit to two subinstances $s^E_1$ and $s^E_2$ when action $x$ is taken. Nevertheless, this is not a serious issue because the following transitions from the child nodes are \emph{independent} with each other. In other words, this can be considered as ``cloning'' the same MDP multiple times, which continue their own transitions thereafter. Based on this observation, for a state $s$ and action $a\in A(s)$, let $S'(s,a)$ be the set of next states. Then the reward of taking $a$ in $s$ is $r(s,a)=\sum_{s'\in S'(s,a)} r(s,a,s')=|S'(s,a)|$. Therefore, for a deterministic policy $\pi$, the value $v^\pi(s)$ of a state $s$ corresponds to the number of search nodes needed to solve the subinstance $s$ following $\pi$, if the discounting factor $\gamma=1$. The goal of the RL agent is to find the optimal policy $\pi^*$ such that the expected (discounted) search tree size is minimized. The optimal action-value function $Q^*(s,a)$ can be expressed recursively based on the following Bellman optimality equation\footnote{From this point on, we will omit the dependency of $S'$ on $s$ and $a$ for brevity.}:
\begin{equation}\label{eq:origBellman}
	Q^*(s,a)=r(s,a) +  \gamma \sum_{S'}{Pr(S' |s, a)
	\sum_{s'\in S'} \min_{a'\in A(s')}
	Q^*(s',a')},
\end{equation}
where $Pr(S'|s,a)$ is the probability of transiting to state set $S'$ if action $a$ is taken in state $s$. If $Q^*$ is known, then the optimal policy is simply to select at each state the action with the minimum $Q^*$ value, i.e. $\pi^*(s)=\arg\min_{a\in A(s)} Q^*(s,a)$. 

The above one-to-many state transitions have already been noticed and handled in \citep{lagoudakis2000algorithm,lagoudakis2001learning}, where RL is applied to learn policies for recursive algorithm selection and choosing branching literals in the \#DPLL procedure for solving SAT problems (with the requirement of finding all solutions). More specifically, Q-learning is used to learn a linearly parameterized function $Q_\mathbf{w}$ as the estimation of $Q^*$. Given a transition $(s, a, S')$ with cost $r(s,a)$, the parameters $\mathbf{w}$ are updated using the following target:
\begin{equation}
	y= r(s,a) + \gamma \sum_{s'\in S'}  \min_{a'\in A(s')} Q_\mathbf{w}(s',a').
\end{equation}

However, the learning mechanisms in \citep{lagoudakis2000algorithm,lagoudakis2001learning} are not suitable for our situation. The key difficulty is that, we do not know $S'$ until the search algorithm backtracks from $s$. Consider again the example in Figure \ref{fig:enum}, we know that $s^E$ has four children at most, but only when backtracking from $s^E$ can we know that only two of them are needed to be explored. This is not an issue for \#DPLL because the algorithm needs to visit all the child nodes eventually. However, in our case, learning must be delayed until backtracking, when the complete transition $(s, a, S')$ and its cost $r(s,a)$ are available so that the target can be computed. This is not desirable because it slows down the learning process, and requires additional engineering efforts to store information for all the pending states.

To make learning faster and more convenient, we intend to achieve bootstrapping using simple transitions $(s,a,s')$ and avoid the need of storing pending states. To this end, we stick to the one-to-one state transition in the typical MDP. More specifically, for state $s$ and action $a$, we consider the multiple child states in $S'$ as samples drawn from the transition dynamics $Pr(s'|s,a)=1/|S'|$ for all $s'\in S'$, where $Pr(s'|s,a)$ is the probability that the environment will transit to $s'$ if $a$ is taken in $s$. In this way, we can use the typical Bellman optimality equation to express an optimal action-value function that are different from that in Equation (\ref{eq:origBellman}):
\begin{equation} \label{eq:qhat}
	\hat{Q}^*(s,a)=  r(s,a,s') + \gamma \sum_{s'}{ Pr(s'|s,a) \min_{a'\in A(s')} \hat{Q}^*(s',a')}.
\end{equation}
An advantage of $\hat{Q}^*$ is that, we can easily apply Q-learning to learn an estimate $\hat{Q}$ of it by bootstrapping according to simple transition $(s,a,s')$ and cost $r(s,a,s')$, since they are immediately known after a child node $s'$ is created. Note that the learned $\hat{Q}$ function is only used to make decision for a given state $s$ when it is visited for the first time, i.e. when the left most branch is created. We denote such action as $a(s)=\arg\min_{a\in A(s)}\hat{Q}^*(s,a)$. For other branches, i.e. those created upon backtracking, the same action is imposed, instead of chosen according to $\hat{Q}$, so that the requirement of backtracking search is satisfied. This is natural for the testing phase, since a fixed $\hat{Q}$ always predicts the same value for given $s$ and $a$. But for training, $\hat{Q}$ keeps changing, hence we need to enforce the same action for transitions to all the following states in $S'$.

Note that the optimal policies $\hat{\pi}^*(s)$ and $\pi^* (s)$ derived from $\hat{Q}^*$ and $Q^*$ are different. In fact, $\hat{\pi}^*(s)$ minimizes the expected cost of reaching a leaf node in the subtree rooted from $s$. This is aligned with the well-known ``fail-first" principle \citep{haralick1980increasing} in designing CSP searching strategy, which suggests to reach the leaf nodes as quick as possible. In Section \ref{sec:experiments}, we will show that this intuition is verified: along with the progress of learning an estimate of $\hat{Q}^*$, the search tree size also decreases though it is not the direct objective to be optimized. In the following section, we will design a parameterized function $\hat{Q}_\mathbf{w}$ to estimate $\hat{Q}^*$ by using a deep neural network.

\subsection{GNN based Representation}
To parameterize $\hat{Q}$, we need to find a way to represent $s$ and $a$. Recall that $s$ is a CSP instances or subinstance which can be described as a constraint network, and $a$ is an unbounded variable. For binary CSP, i.e. the arity $|scp(c)|=2$ for all constraint $c\in\mathcal{C}$, the underlying constraint network can be viewed as a graph with the variables being vertices and constraints being edges. Such a graph can be naturally represented by GNN frameworks. Essentially, a GNN learns a vector representation, or embedding, for each vertex in a given graph by iteratively performing embedding aggregation among neighboring vertices \citep{xu2018powerful}. But in general, the arity of CSP constraints could be larger than 2, meaning that the underlying structure is a \emph{hypergraph}, with the constraints being hyperedges. To effectively represent the constraint network, below we design a GNN variant that learns embeddings for both the vertices and hyperedges. 

Given a constraint network $P$, let $\mathcal{N}_c(j)=\{i|x_i\in scp(c_j) \}$ be the indexes of variables that are in the scope of a constraint $c_j$, and $\mathcal{N}_v(i)=\{j|x_i\in scp(c_j) \}$ be the indexes of constraints where a variable $x_i$ is involved in. The current status of variables and constraints are characterized by raw feature vectors $X_i$ and $C_j$, each with dimension $p_v$ and $p_c$. Our GNN computes a $p$-dimensional embedding $\mu_i$ and $\nu_j$ for each variable $x_i\in\mathcal{X}$ and constraint $c_j\in\mathcal{C}$.\footnote{Here we use the same embedding dimension for variables and constraints for simplicity; but in general they could have different dimensions.} These embeddings are first initialized by linearly transforming the respective raw feature vectors, i.e. $\mu_i^{(0)}= X_i \mathbf{w}_v$ and $\nu_i^{(0)}= C_i \mathbf{w}_c$, where $\mathbf{w}_v \in \mathbb{R}^{p_v\times p}$ and $\mathbf{w}_c \in \mathbb{R}^{p_c\times p}$ are learnable parameters. Then we update these embeddings by performing $K$ iterations of embedding aggregation operations among the variables and constraints, based on the underlying hypergraph structure. More specifically, in each iteration $k=1,...,K$, we perform the following steps:
\begin{itemize}
	\item Embedding of each constraint $c_j$ is first updated by aggregating embeddings from the related variables in $\mathcal{N}_c(j)$. More specifically, we use element-wise summation as the aggregation function, the result of which is fed into a Multilayer Perceptron (MLP) $\text{MLP}_v$ to get the updated embedding of $c_j$, along with its embedding in the previous round and its raw feature vector. This procedure is shown as follows:
	\begin{equation} \label{eq:var_to_constr}
		\nu^{(k)}_j \gets \text{MLP}_v\left[ \sum\nolimits_{i\in \mathcal{N}_c(j)} \mu^{(k-1)}_i:\nu_j^{(k-1)} : C_i \right],
	\end{equation}
	where $[\cdot:\cdot]$ is the concatenation operator. The input and output dimension of $\text{MLP}_v$ are $2p+p_c$ and $p$, respectively.
	\item Embedding of each variable $x_i$ is updated by aggregating embeddings of the related constraints in $\mathcal{N}_v(i)$, based on similar procedure shown below:
	\begin{equation} \label{eq:constr_to_var}
		\mu^{(k)}_i \gets \text{MLP}_c\left[ \sum\nolimits_{j\in \mathcal{N}_v(i)} \nu^{(k)}_i: \mu_i^{(k-1)} : X_i\right].
	\end{equation}
	The input and output dimension of $\text{MLP}_c$ are $2p+p_v$ and $p$, respectively.
\end{itemize}

To parameterize $\hat{Q}_\mathbf{w}$, we represent the current state $s$ by performing graph-level pooling using element-wise summation of all the variable embeddings after iteration $K$, i.e. $\sum_{i=1}^n \mu^{(K)}_i$, similar to \citep{khalil2017learning}. Then we concatenate the embedding representations of the graph and corresponding action $a$, and feed it into another MLP to get $\hat{Q}_\mathbf{w}$ as follows:
\begin{equation} \label{eq:Q_s_a}
	\hat{Q}_\mathbf{w}(s,a) = \text{MLP}_q\left[
	\sum\nolimits_{i=1}^n \mu^{(K)}_i : \mu^{(K)}_a
	\right],
\end{equation}
where $\text{MLP}_q$ has input and output dimension of $2p$ and 1, respectively.

The raw features of variables and constraints in a state $s$ are summarized below. For each variable $x_i$, we use its current domain size $|d(x_i,s)|$ and a binary indicator $b(x_i,s)$ specifying whether it is bounded as the raw features, hence the vector dimension $p_v=2$. For each constraint $c_j$, its raw feature vector contains: 1) the number of unbounded variables $ub(c_j,s)$, and 2) the current constraint tightness $1-|rel(c_j,s)|/D(c_j,s)$, where $|rel(c_j,s)|$ is the number of currently allowed tuples and $D(c_j,s)=\prod_{x_i\in scp(c_j)}|d(x_i,s)|$ is the product of current domain sizes of the involved variables. The dimension of constraint feature vector is $p_c=2$.

\begin{figure}
	\centering
	\includegraphics[width=\linewidth]{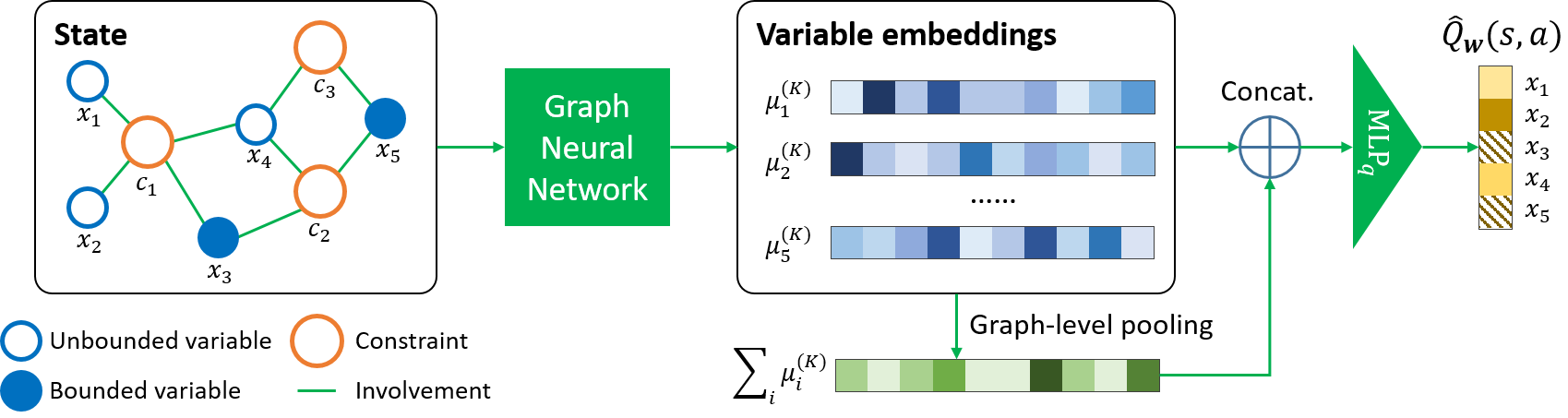}
	\caption{Illustration of the Q network. The outputs of $x_3$ and $x_5$ are masked out since they are bounded.}
	\label{fig:QNet}
\end{figure}

The computation process of the Q network is visualized in Figure \ref{fig:QNet}. In the implementation, to make full use of the parallel computing power of GPU, we collect the raw features of all the variables and constraints into two matrices $\mathbf{X} \in \mathbb{R}^{n\times p_v}$ and $\mathbf{C} \in \mathbb{R}^{e \times p_c}$ where each row is the corresponding raw feature $X_i$ and $C_j$, and feed them into the network. In this case, the computations in Equations (\ref{eq:var_to_constr})-(\ref{eq:Q_s_a}) can be done in parallel for all the variables and constraints, in the form of matrix calculations. To be more specific, let $\boldsymbol\mu^{(k)} \in \mathbb{R}^{n \times p}$ and $\boldsymbol\nu^{(k)} \in \mathbb{R}^{e \times p}$ be the embeddings of all the variables and constraints after iteration $k$. Then the embedding update in Equations (\ref{eq:var_to_constr}) and (\ref{eq:constr_to_var}) can be written as:
\begin{equation}
    \boldsymbol\nu^{(k)}\gets \text{MLP}_v\left[ \mathbf{A}\boldsymbol\mu^{(k-1)} : \boldsymbol\nu^{(k-1)} : \mathbf{C} \right], 
\end{equation}
\begin{equation}
    \boldsymbol\mu^{(k)}\gets \text{MLP}_c\left[ \mathbf{A}^{\top}\boldsymbol\nu^{(k)} : \boldsymbol\mu^{(k-1)} : \mathbf{X} \right], 
\end{equation}
where $\mathbf{A}$ is a $e \times n$ binary matrix representing the adjacency relations between variables and constraints, and $\mathbf{A}_{ji}$ is 1 if $x_i\in scp(c_j)$ and 0 otherwise. Note that we implement $\mathbf{A}$ as a sparse matrix, and the number of non-zero values in $\mathbf{A}$ is $\mathcal{S} = \sum_{j=1}^e scp(c_j)$. The matrix form of Equation (\ref{eq:Q_s_a}) is a bit different from its definition. Specifically, we compute the Q-values for all the variables, and mask out the results of bounded variables (i.e. set the Q-value to $\infty$) such that only legal actions in $A(s)$ will be selected. In this case, Equation (\ref{eq:Q_s_a}) can be written as:
\begin{equation}
    \hat{\mathbf{Q}}_\mathbf{w}(s)=\text{MLP}_q\left[\mathbf{I}\boldsymbol\mu^{(K)} : \boldsymbol\mu^{(K)} \right],
\end{equation}
where $\mathbf{I}$ is a $n\times n$ all-one matrix.

Here we briefly analyze the time complexity of the Q-network, for which the major computation is on matrix multiplications. To simplify the analysis, we consider the embedding dimension $p$ and raw feature dimensions $p_v$ and $p_c$ as constant. Since $\mathbf{A}$ is sparse, the complexity of Equation (\ref{eq:var_to_constr}) is $O\left(\mathcal{S}p+ep(2p+p_c)\right)=O(\mathcal{S}+e)$, where $O(\mathcal{S}p)$ is for sparse-dense matrix multiplication and $O(ep(2p+p_c))$ is for dense matrix multiplications in $\text{MLP}_v$. Similarly, the computation of Equation (\ref{eq:constr_to_var}) is in $O(\mathcal{S}+n)$, and the complexity of $K$ iterations is $O\left(K(\mathcal{S}+n+e)\right)$. Computation of the Q-values in Equation (\ref{eq:Q_s_a}) is in $O(n^2p+n(2p))=O(n^2)$, and hence the overall time complexity of the Q-network is $O\left(K(\mathcal{S}+n+e)+n^2\right)$.

The above representation scheme inherits the nice property of GNN, i.e. the same model and set of parameters can process instances with arbitrary size and constraint arity, because all the trainable parameters are shared across all its inputs (e.g. $\text{MLP}_v$ is \emph{shared} across all constraints). This effectively enables generalizing models trained on small instances to larger ones. In the next section, we describe our algorithm for training the RL agent.

\subsection{Training Algorithm}
Our training algorithm is designed based on Double Deep Q-Network (DDQN) \citep{van2016deep}. It maintains two networks, i.e. the online network $\hat{Q}_\mathbf{w}$ and target network $\hat{Q}_\mathbf{\bar{w}}$. More specifically, $\hat{Q}_\mathbf{\bar{w}}$ is a periodical copy of $\hat{Q}_\mathbf{w}$. At each state $s$, the RL agent selects an action $a_t$ using the $\epsilon$-greedy strategy according to the online network, and the experience $(s,a,s',r,T)$ is added to an experience reply buffer $\mathcal{R}$ with memory size $\mathcal{M}$, where $T=T(s')$ is a binary indicator of whether $s'$ is a terminal state. Then a mini-batch of $\mathcal{B}$ transitions is sampled from $\mathcal{R}$ to update the parameters of the online network $\hat{Q}_\mathbf{w}$ by performing gradient decent to minimize the squared loss between $\hat{Q}_\mathbf{w}(s,a)$ and the following target:
\begin{equation} \label{eq:DDQNTarget}
	y = r + \gamma \hat{Q}_\mathbf{\bar{w}}\left(s',\underset{a'\in A(s')}{\arg\min}\hat{Q}_\mathbf{w}(s',a')\right).
\end{equation}
Note that the above target computation is only applicable when the state $s'$ is non-terminal. For the terminal ones, the target is simply $y=r$.

\begin{algorithm}[!t]
	\caption{DDQN for learning variable ordering heuristic}
	\label{alg:training}
	\begin{algorithmic}
		\State Initialize the experience replay to capacity $\mathcal{M}$
		\For {episode $e=1$ to $\mathcal{N}$}
		\State Draw a CSP instance $P\sim \mathbb{D}$
		\State $\mathcal{T}\gets 0$
		\While {$\mathcal{T} < \mathcal{T}_{max} $ and $P$ is not solved yet} 
		\State Observe state $s$, 
		\If{s has been visited}
		\State Choose $a=a(s)$
		\Else
		\State Choose $a$ as follows:
		\State $a(s) = \left\{ 
		\begin{array}{ll}
			\text{randomly choose from } A(s) & \text{w.p. } \epsilon \\
			\arg\min_{a\in A(s)} \hat{Q}_\mathbf{w}(s,a) & \text{otherwise}
		\end{array}
		\right. $
		\EndIf
		\State Execute $a$, observe $s'$ and $T(s')$
		\State Store $(s,a,s',1, T(s'))$ in $\mathcal{R}$
		\State Randomly sample a minibatch from $\mathcal{R}$
		\State For each sampled experience, compute the target:
		\State $y = \left\{
		\begin{array}{ll}
			r & \text{If } s' \text{ is terminal} \\
			\text{use Equation  (\ref{eq:DDQNTarget})} & \text{otherwise}
		\end{array}
		\right. $
		\State Perform a gradient descent step to update $\mathbf{w}$
		\State $\mathcal{T}\gets \mathcal{T}+1$
		\If {$P$ is solved}
		\State \textbf{Break while}
		\EndIf
		\EndWhile
		\State For every $e_u$ episodes, set $\mathbf{\bar{w}} = \mathbf{w}$
		\EndFor
	\end{algorithmic}
\end{algorithm}

Our training algorithm is shown in Algorithm  \ref{alg:training}. The agent is trained for $\mathcal{N}$ episodes, during each of them the agent tries to solve a CSP instance drawn from the distribution $\mathbb{D}$. Due to the intractability of CSP, it is possible that solving an instance requires a very large number of steps, i.e. state transitions, especially in the beginning stage of learning when $\epsilon$ is large and the quality of policy is low. Though we can let the agent finish solving an instance, this is not desirable because the agent's experience may be limited to a small number of instances for a long time. Therefore, a special design here is that we set a cutoff limit of $\mathcal{T}_{max}$ steps (equivalent to the maximum number of search nodes) to limit the effort spent by the agent on one instance. Note that the terminal indicator $T(s')$ of an experience is true only when $s'$ corresponds to  a leaf node. For those $s'$ terminated due to reaching $\mathcal{T}_{max}$, $T(s')$ is still false. This is to ensure that the target is correctly computed: the actual cost of a state $s'$ terminated by $\mathcal{T}_{max}$ is not 0, since more nodes under it need to be visited to solve the subinstance in $s'$. This corresponds to the partial-episode bootstrapping method in \citep{pardo2018time}. In the experiments, we will show that even a large number of instances hit the cutoff limit in the early stage, the agent can still learn high-quality policy that solves all testing instances.

\section{Computational Experiments} \label{sec:experiments}
In this section, we conduct a series of experiments to test the proposed approach. We first introduce the setup of our experiments, then present the training and testing results on small-sized instances, and finally report the generalization performance on larger instances, as well as some analysis of the execution time.

\subsection{Experimental Setup}

\textbf{Instance generation.} The CSP instance used in our experiments are generated using the well-known and widely used random CSP generator, model RB \citep{xu2007random}. It takes 5 parameters $<m,n,\alpha,\beta,\rho>$ as input to generate a CSP instance, the meanings of which are listed below:
\begin{itemize}
	\item $m\geq 2$ is the arity of each constraint;
	\item $n\geq 2$ is the number of variables;
	\item $\alpha>0$ specifies $d$, which is the domain size of each variable, and $d=n^\alpha$;
	\item $\beta>0$ specifies $e$, which is the number of constraints, and $e=\beta\cdot n\cdot \ln n$;
	\item $\rho\in(0,1)$ specifies the constraint tightness, and $\rho\cdot d^k$ is the number of disallowed tuples for each constraint.
\end{itemize}
Each unique combination of the above parameters specifies a class of CSP instances, which can be considered as the distribution $\mathbb{D}$. The CSP classes used in our experiments are all situated at the \emph{phase transition} thresholds, which are combinations of parameters that result in the \emph{hardest} instances. A nice theoretical property of model RB that makes it more preferable than other random CSP models is that, it can guarantee exact phase transitions and instance hardness at the threshold \citep{xu2007random}. We test our approach for two types of distributions with binary and 3-ary constraints, denoted as $\mathbb{D}_1(n)=<2,n,0.7,3,0.21>$ and $\mathbb{D}_2(n)=<3,n,0.7,2.5,0.24>$, respectively. With different $n$, we have CSP classes with different sizes. In our experiments, we choose $n$ from $\{15,25,30,40\}$ and $\{10,15,20,25\}$ for $\mathbb{D}_1(n)$ and $\mathbb{D}_2(n)$, respectively, since higher constraint arity generally leads to harder instances.

\textbf{Implementation details.} For our GNN model, we instantiate it by setting the embedding dimension $p=128$, and all MLPs have $L=3$ layers with hidden dimension 128 and rectified linear units as activation function. The embeddings are updated for $K=5$ iterations. We train the RL agent for $\mathcal{N}=1000$ episode, i.e. solving 1000 instances drawn from distribution $\mathbb{D}$, with the cutoff step limit $\mathcal{T}_{max}=10000$. During training, another 200 instances drawn from $\mathbb{D}$ are used to validate performance of the agent's policy. The discount factor $\gamma$ is set to 0.99. For exploration, the value of $\epsilon$ is set to 1 in the beginning, and linearly annealed to 0.05 in the first 20000 steps. We use the Adam optimizer to train the neural network, with a constant learning rate $\eta=0.00005$ and mini-batch size $\mathcal{B}=128$. The size of experience replay is $\mathcal{M}=0.1\text{M}$. The frequency of updating the target network is $e_u=100$.  For testing, we set a cutoff limit of $5\times 10^5$ search nodes for our policies and all baseline heuristics. The neural architecture and  hyperparameters are empirically tuned on small sized instances. More details will be given in Section \ref{sec:impact}. Our approach is implemented in C++ on top of the source code of Google OR-Tools\footnote{https://github.com/google/or-tools. Note that our implementation is based on the original CP solver, instead of the CP-SAT solver.}, a state-of-the-art CSP solver, which employs the binary branching strategy. The GNN architecture and training algorithm is implemented based on the source code\footnote{https://github.com/Hanjun-Dai/graph\_comb\_opt.} of \citep{khalil2017learning}. The hardware we used is a workstation with Intel Core i9-9920x CPU and one NVIDIA RTX 2080Ti GPU (11GB memory). Our code is publicly available at \url{https://github.com/songwenas12/csp-drl}.

\textbf{Baselines.} We compare the trained policies with four classic hand-crafted variable ordering heuristics that are representative and commonly used in many CSP solvers. The first two are embedded in the underlying solver OR-Tools (other embedded heuristics are rather uncompetitive in the experiments hence ignored here):
\begin{itemize}
	\item \textsf{MinDom} \citep{haralick1980increasing}, which selects the variable with the minimum current domain size (Dom). Despite its simplicity, this heuristic is effective and popular in practice, and is implemented in almost all solvers. 
	\item Impact-based heuristic (\textsf{Impact}) \citep{refalo2004impact}, which selects the variable that can lead to an assignment with the maximum reduction of search space, i.e. impact. 
	\textsf{Impact} is considered as one of the state-of-the-art variable ordering heuristics \citep{li2016improving}, and is implemented as the default search strategy in OR-Tools. Here we use its default configuration where the impact of a variable is measured by summing the impact of each value in its current domain, and the decision logic is to select the variable with the maximum impact and the value with the minimum impact.	 
\end{itemize}

Note that even with its embedded heuristics, OR-tools is already a very competitive solver and has won the MiniZinc Challenge \citep{stuckey2014minizinc} several times. Nevertheless, we implement the below two heuristics for comparison:
\begin{itemize}
	\item \textsf{Dom/Ddeg} \citep{bessiere1996mac}, which improves \textsf{MinDom} by taking the dynamic degree (Ddeg) of a variable into account, and selects the variable with the minimum ratio between Dom and Ddeg. To compute the Ddeg of a variable $x_i$, this heuristic first identifies the set of constraints involving $x_i$, i.e. $\mathcal{C}(x_i)=\{c_j\in \mathcal{C}|x_i\in scp(c_j)  \}$; then it removes from $\mathcal{C}(x_i)$ those constraints that involve no unbounded variables, and uses the cardinality of $\mathcal{C}(x_i)$ as the Ddeg value of $x_i$. 
	\item \textsf{Dom/Tdeg} \citep{li2016improving}, which is a recently proposed heuristic that incorporates constraint tightness that is also used in our method as a feature. It replaces the Ddeg in \textsf{Dom/Ddeg} by the dynamic tightness degree (Tdeg), defined as the summation of tightness of constraints in $\mathcal{C}(x_i)$, instead of the cardinality as in Ddeg.
\end{itemize}

Except \textsf{Impact} which has its own value ordering heuristic, we apply lexicographical ordering for our approach (denoted as \textsf{DRL}) and the other baselines \textsf{MinDom}, \textsf{Dom/Ddeg} and \textsf{Dom/Tdeg} to select the next value.

\subsection{Training and Testing on Small and Medium Sized Instances}
In this section, we discuss the performance of the RL agent during training. More specifically, we train the agent on small and medium sized binary and 3-ary instance distributions including $\mathbb{D}_1(15)$, $\mathbb{D}_1(25)$,  $\mathbb{D}_2(10)$ and $\mathbb{D}_2(20)$, respectively. We use two measures to evaluate the agent's policies, including 1) the number of search nodes, which is the objective we try to minimize, and 2) the number of failures (i.e. dead-ends), which is a measure of the ability to ``fail first" \citep{beck2004trying} and hence reflects the performance with respect to the objective we defined in Equation (\ref{eq:qhat}).

\begin{figure}
	\centering
	\begin{subfigure}[b]{0.47\textwidth}
		\centering
		\includegraphics[width=\textwidth]{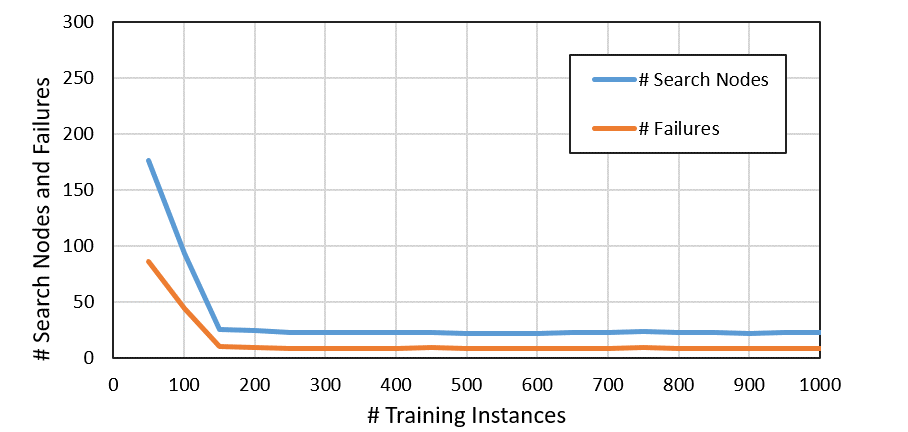}
		\caption{On distribution $\mathbb{D}_1(15)$}
		\label{fig:training_curve1}
	\end{subfigure}
	\begin{subfigure}[b]{0.47\textwidth}
		\centering
		\includegraphics[width=\textwidth]{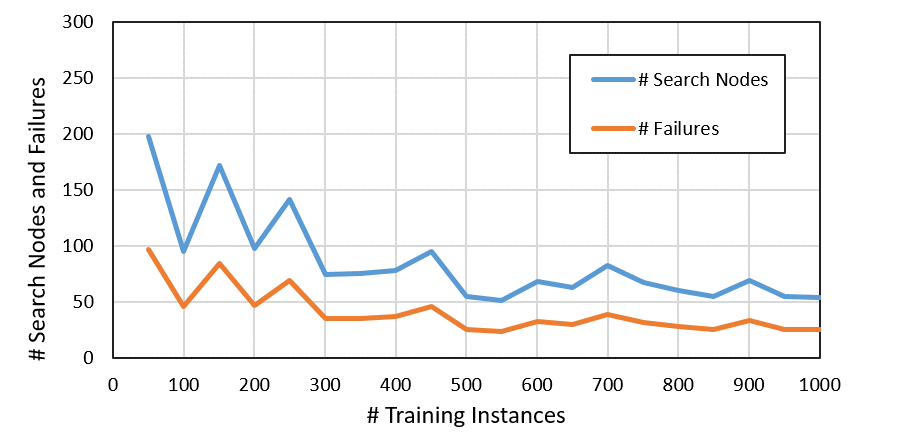}
		\caption{On distribution $\mathbb{D}_2(10)$}
		\label{fig:training_curve2}
	\end{subfigure}
	\begin{subfigure}[b]{0.47\textwidth}
		\centering
		\includegraphics[width=\textwidth]{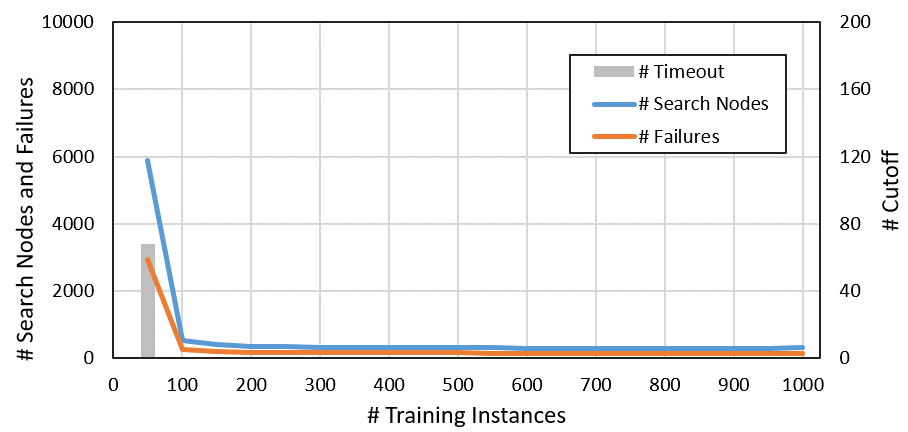}
		\caption{On distribution $\mathbb{D}_1(25)$}
		\label{fig:training_curve3}
	\end{subfigure}
	\begin{subfigure}[b]{0.47\textwidth}
		\centering
		\includegraphics[width=\textwidth]{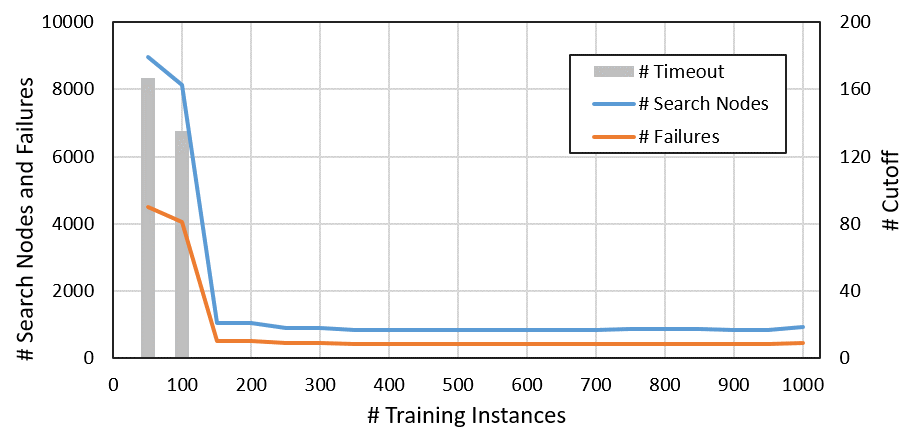}
		\caption{On distribution $\mathbb{D}_2(15)$}
		\label{fig:training_curve4}
	\end{subfigure}
	\caption{Training Performance (Evaluated every 50 training instances).}
	\label{fig:training_curves}
\end{figure}

In Figures \ref{fig:training_curve1} and \ref{fig:training_curve2}, we plot the agent's training performance on the two smallest distributions, i.e. $\mathbb{D}_1(15)$ and $\mathbb{D}_2(10)$, with respect to the average values of the above two measures on the 200 validation instances. For these two distributions, all instances are successfully solved within the 10000 cutoff step limit. As can be observed, the agent's performance significantly improved during training. For example, on $\mathbb{D}_1(15)$, the agent needs to visit over 150 search nodes with more than 70 times of failure on average in the beginning stage to solve an instance. With the increase of training episodes (i.e. the number of training instances), the two measures significantly drops to 22-23 search nodes and 8-9 failures. Similar trends also exist in Figure \ref{fig:training_curve2}. However, the training process on $\mathbb{D}_1(10)$ is more fluctuated, and requires more episodes to converge to a policy with 50-100 search nodes and 30-50 failures. This indicates that though having smaller number of variables, solving instances drawn from $\mathbb{D}_2(10)$ is harder than solving those from $\mathbb{D}_2(15)$, in terms of both learning and solving.

Figures \ref{fig:training_curve3} and \ref{fig:training_curve4} show the training curves on medium sized distributions $\mathbb{D}_1(25)$ and $\mathbb{D}_2(15)$. Clearly, instances drawn from these two distributions are much harder to solve. Despite the large number of search nodes and failures, the agent even cannot solve many validation instances within the cutoff limit of 10000 nodes in the early stage. For example, on the hardest distribution $\mathbb{D}_2(15)$, there are still 67.5\% (135 out of 200) validation instances cannot be solved after 100 episodes of training. Nevertheless, the agent can still drastically improve its performance. For example, on $\mathbb{D}_2(15)$, the number of search nodes significantly drops from nearly 9000 to about 830, and all validation instances are successfully solved. Note that on the same scale, the cures in Figures \ref{fig:training_curve3} and \ref{fig:training_curve4} are more fluctuated than their counterparts in Figures \ref{fig:training_curve1} and \ref{fig:training_curve2}.

Figure \ref{fig:training_curves} demonstrates the effectiveness of our design in several aspects. Firstly, for the learning tasks on distributions with different features and hardness, the agent is able to learn a variable ordering policy from scratch by itself, without the need of supervision. Secondly, though the agent is optimizing an alternative objective defined in Equation (\ref{eq:qhat}), with the progress of learning, the search tree size also decreases with almost identical trend. Finally, the large number of cutoff instances in the early stages does not affect the training process to converge to high-quality policies.

\begin{table}[!t] \small
	\centering
	\caption{Test results on the same distribution used in training}
	\label{tb:test_small}
	\begin{tabular}{ccrrrr}
		\toprule
		\multirow{2}{*}{$\mathbb{D}$} & \multirow{2}{*}{Heuristic} & \multicolumn{2}{c}{\# Search Nodes} & \multicolumn{2}{c}{\# Failures} \\
		\cmidrule(lr){3-4}
		\cmidrule(lr){5-6}
		& & Average  & Reduction  & Average & Reduction \\ \midrule
		\multirow{5}{*}{\rotatebox[origin=c]{90}{$\mathbb{D}_1(15)$}} & \textsf{DRL}      & \textBF{22.21}   & - & \textBF{8.60}            & -    \\
		& \textsf{Dom/Tdeg} & \textBF{22.81}        & -     & \textBF{8.91}      & -             \\
		& \textsf{Dom/Ddeg} & 23.05        & 3.64\%     & 9.02      & 4.66\%             \\
		& \textsf{MinDom}   & 33.57         & 33.85\%              & 14.15      & 39.22\%            \\
		& \textsf{Impact}   & 272.51       & 91.85\%              & 134.99     & 93.63\%  \\ \midrule
		\multirow{5}{*}{\rotatebox[origin=c]{90}{$\mathbb{D}_2(10)$}} & \textsf{DRL}      & \textBF{56.65}        & -                    & \textBF{26.60}       & -     \\
		& \textsf{Dom/Tdeg} & \textBF{57.82}        & -     & \textBF{27.12}      & -             \\
		& \textsf{Dom/Ddeg} & 59.98        & 5.55\%                 & 28.27       & 5.91\%                \\
		& \textsf{MinDom}   & 100.46         & 43.61\%                & 48.40      & 45.04\%              \\
		& \textsf{Impact}   & 514.09       & 88.98\%                & 267.75     & 90.07\%   \\ \midrule
		\multirow{5}{*}{\rotatebox[origin=c]{90}{$\mathbb{D}_1(25)$}} & \textsf{DRL}      & \textBF{291.30}        & -                    & \textBF{141.92}       & -     \\
		& \textsf{Dom/Tdeg} & 320.19        & 9.02\%     & 156.26      & 9.18\%             \\
		& \textsf{Dom/Ddeg} & 347.78  & 16.24\%                 & 170.06       & 16.54\%                \\
		& \textsf{MinDom}   & 799.54         & 63.57\%                & 395.82      & 64.14\%              \\
		& \textsf{Impact}   & 69885.05       & 99.58\%                & 37526.49     & 99.62\%   \\ \midrule
		\multirow{5}{*}{\rotatebox[origin=c]{90}{$\mathbb{D}_2(15)$}} & \textsf{DRL}      & \textBF{972.45}        & -                    & \textBF{483.58}       & -     \\
		& \textsf{Dom/Tdeg} & 1084.81        & 10.36\%     & 539.80      & 10.41\%             \\
		& \textsf{Dom/Ddeg} & 1143.85  & 14.98\%                 & 569.25       & 15.05\%                \\
		& \textsf{MinDom}   & 2537.24         & 61.67\%   & 1265.90      & 61.80\%    \\
		& \textsf{Impact}   & 48941.93       & 98.01\%                & 26267.43     & 98.16\%   \\
		\bottomrule          
	\end{tabular}
\end{table}

We then evaluate the quality of the trained policies by comparing them with the baseline heuristics on four test sets, each with 500 instances drawn from the corresponding distributions used in training.  We use the best policy (with respect to the number of search nodes) to obtain the results of our approach \textsf{DRL}. In these experiments, all the testing instances are successfully solved by all heuristics within the cutoff limit of $5\times 10^5$ search nodes. The results are summarized in Table \ref{tb:test_small}, where the column ``Average" is the mean number of search nodes (failures) for each heuristic on the 500 testing instances, and the column ``Reduction" is the percentage reduction in the average number of search nodes (failures) that our policies achieved compared with each baseline. We can observe that for all the four distributions, the policies trained by our approach outperform the two baseline heuristics embedded in OR-Tools, i.e. \textsf{MinDom} and \textsf{Impact}, by a large margin. The reason why \textsf{Impact} performs not quite well is probably because random CSP instances do not have very strong structure \citep{correia2008efficiency}. For the other two baselines, our method consistently outperforms \textsf{Dom/Ddeg} on all the four distributions. In terms of \textsf{Dom/Tdeg}, our method performs on par with it on the two small distributions $\mathbb{D}_1(15)$ and $\mathbb{D}_2(10)$, since the difference between \textsf{DRL} and \textsf{Dom/Tdeg} is not statistically significant\footnote{We use the Wilcoxon signed-rank test with $p<0.001$ for all the statistical significant tests in this paper.}. However, on the two larger and harder distributions $\mathbb{D}_1(25)$ and $\mathbb{D}_2(15)$, \textsf{DRL} significantly outperforms \textsf{Dom/Tdeg} with around 10\% reduction in the number of nodes and failures, and is statistically significant (for \# nodes and failures, $p$ value is $2.94\times 10^{-10}$ and $3.52\times 10^{-10}$ on $\mathbb{D}_1(25)$, and $5.22\times 10^{-12}$ and $4.87\times 10^{-12}$ on $\mathbb{D}_2(15)$). This key observation indicates that our method could be more effective on harder problems. A possible explanation is that, the search tree for harder problems are larger, hence the space for possible improvement is also larger.

\subsection{Generalizing to Larger Instances}
As mentioned previously, our GNN based representation enables generalizing the trained models to larger instances that have never been seen by the agent during training. In this section, we conduct experiments to evaluate the generalization performance. Specifically, for distributions $\mathbb{D}_1(n)$, we run the policy trained on $\mathbb{D}_1(25)$ on larger distributions with $n\in\{30,40\}$; for the distributions of 3-ary CSP $\mathbb{D}_2(n)$, the policy trained on $\mathbb{D}_2(15)$ is evaluated on distributions with $n\in\{20,25\}$. Similar as in the previous section, we randomly sample 500 instances from each distribution as the test set. In Tables \ref{tb:gen_1} and \ref{tb:gen_2}, we summarize the results on binary and 3-ary CSP distributions $\mathbb{D}_1(n)$ and $\mathbb{D}_2(n)$, respectively. In these tables, values in the columns ``Average" are computed using all results on the 500 testing instances, but those in columns ``Reduction" are computed based on the instances that are solved by both \textsf{DRL} and the comparing baseline. The columns ``\# Cutoff" show the number of instances that are not solved by each heuristic within the cutoff limit of $5\times 10^{5}$ nodes. Note that for those experiments that a majority ($\geq 250$) of instances are cut off, we do not report the reduction values.

\begin{table}[!t] \small
	\centering
	\caption{Generalization results on binary CSP distributions}
	\label{tb:gen_1}
	\begin{threeparttable}
	\begin{tabular}{ccrrrrr}
		\toprule
		\multirow{2}{*}{$\mathbb{D}$} & \multirow{2}{*}{Heuristic} & \multicolumn{2}{c}{\# Search Nodes} & \multicolumn{2}{c}{\# Failures} & \multirow{2}{*}{\# Cutoff} \\ 
		\cmidrule(lr){3-4}
		\cmidrule(lr){5-6}
		&                            & Average          & Reduction      & Average  & Reduction   &            \\ \midrule
		\multirow{5}{*}{\rotatebox[origin=c]{90}{$\mathbb{D}_1(30)$}}       & \textsf{DRL($\mathbb{D}_1(25)$)}                        & \textBF{1237.93}       & -                   & \textBF{614.44}      & -                & -          \\
		& \textsf{Dom/Tdeg}   & 1350.92       & 8.36\%                & 670.93     & 8.42\%             & -          \\
		& \textsf{Dom/Ddeg}   & 1523.92       & 18.77\%                & 757.38      & 18.87\%             & -          \\
		& \textsf{MinDom}                     & 4160.78       & 70.25\%     & 2075.69     & 70.40\%            & -          \\
		& \textsf{Impact}                     & 318862.66      & -               & 171029.08    & -            & 255        \\ \midrule
		\multirow{5}{*}{\rotatebox[origin=c]{90}{$\mathbb{D}_1(40)$}}       & \textsf{DRL($\mathbb{D}_1(25)$)} & \textBF{23684.04}      & -                   & \textBF{11836.26}    & -                & -          \\
		& \textsf{Dom/Tdeg}       & 26861.89      & 11.83\%                 & 13425.07     & 11.83\%              & -          \\
		& \textsf{Dom/Ddeg}       & 31807.81      & 25.54\%                 & 15897.96     & 25.55\%              & -          \\
		& \textsf{MinDom}                     & 136405.52        & 81.38\%               & 68196.94    & 81.39\%            & 13         \\
		& \textsf{Impact}                     & 491872.80      & -               & 263666.30    & -            & 492    \\   \bottomrule
	\end{tabular}
	\begin{tablenotes}
		\footnotesize
		\item Note: Reduction values are computed using instances solved by both \textsf{DRL} and the corresponding baseline method.
	\end{tablenotes}
	\end{threeparttable}
\end{table}

\begin{table}[!t] \small
	\centering
	\caption{Generalization results on 3-ary CSP distributions}
	\label{tb:gen_2}
	\begin{threeparttable}
	\begin{tabular}{ccrrrrr}
		\toprule
		\multirow{2}{*}{$\mathbb{D}$} & \multirow{2}{*}{Heuristic} & \multicolumn{2}{c}{\# Search Nodes} & \multicolumn{2}{c}{\# Failures} & \multirow{2}{*}{\# Cutoff} \\ 
		\cmidrule(lr){3-4}
		\cmidrule(lr){5-6}
		&                            & Average          & Reduction      & Average     & Reduction   &            \\ \midrule
		\multirow{5}{*}{\rotatebox[origin=c]{90}{$\mathbb{D}_2(20)$}}       & \textsf{DRL($\mathbb{D}_2(15)$)}                        & \textBF{17421.14}       & -                   & \textBF{8707.31}      & -                & -          \\
		& \textsf{Dom/Tdeg}     & 20598.00       & 15.42\%                & 10295.76      & 15.43\%             & -          \\
		& \textsf{Dom/Ddeg}     & 22113.32       & 21.22\%                & 11053.37      & 21.22\%             & -          \\
		& \textsf{MinDom}                     & 73781.04       & 75.46\%               & 36887.05     & 75.46\%            & 3          \\
		& \textsf{Impact}                     & 462299.49      & -               & 248075.41    & -            & 436        \\ \midrule
		\multirow{5}{*}{\rotatebox[origin=c]{90}{$\mathbb{D}_2(25)$}}       & \textsf{DRL($\mathbb{D}_2(15)$)}                        & \textBF{287406.14}      & -                   & \textBF{143701.41}    & -                & \textBF{153}          \\
		& \textsf{Dom/Tdeg}     & 322161.17       & 18.74\%                & 161079.64      & 18.74\%             & 191          \\
		& \textsf{Dom/Ddeg}                   & 330373.63      & 22.08\%                 & 165186.04     & 22.08\%              & 203          \\
		& \textsf{MinDom}                     & 438152.80        & -               & 219079.34    & -            & 391         \\
		& \textsf{Impact}                     & 499595.53      & -               & 268061.19    & -      & 498    \\   \bottomrule
	\end{tabular}
	\begin{tablenotes}
		\footnotesize
		\item Note: Reduction values are computed using instances solved by both \textsf{DRL} and the corresponding baseline method.
	\end{tablenotes}
	\end{threeparttable}
\end{table}

In Table \ref{tb:gen_1}, we can observe that with the increase of $n$, the hardness of solving the instances grows rapidly for all heuristics, which shows the exponential complexity of CSP. For all experiments in this table, our policy trained on $\mathbb{D}_1(25)$ still significantly outperforms all the baselines, with even larger reduction compared with the corresponding results in Table \ref{tb:test_small}. Moreover, the reduction tends to be more significant on instances with larger $n$ which are harder to solve. Our conjecture for this behavior is that, the performance of the trained policy degrades as the increase of problem size which is common for existing deep learning based approaches, e.g. \citep{khalil2017learning,kool2018attention}. However, for larger instances, the spaces that the learned policy can improve over classic heuristics also become larger, which surpass the effect of performance degradation. For Table \ref{tb:gen_2} which summarizes the results for 3-ary distributions, we can make almost the same observations as those for Table \ref{tb:gen_1}. Moreover, we notice that the reductions over the baseline heuristics in these experiments on 3-ary distributions are more prominent than those on binary distributions, especially for the most competitive baseline \textsf{Dom/Tdeg}.

\begin{figure}[!t]
	\centering
	\begin{subfigure}[b]{0.6\textwidth}
		\centering
		\includegraphics[width=\textwidth]{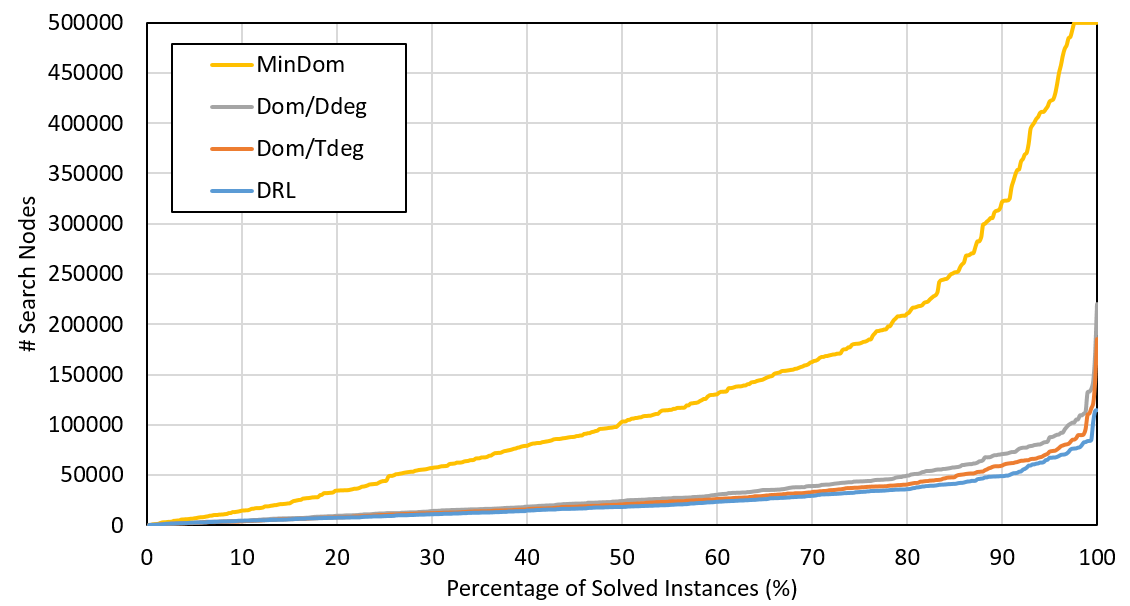}
		\caption{On distribution $\mathbb{D}_1(40)$}
		\label{fig:cactus1}
	\end{subfigure}
	\begin{subfigure}[b]{0.6\textwidth}
		\centering
		\includegraphics[width=\textwidth]{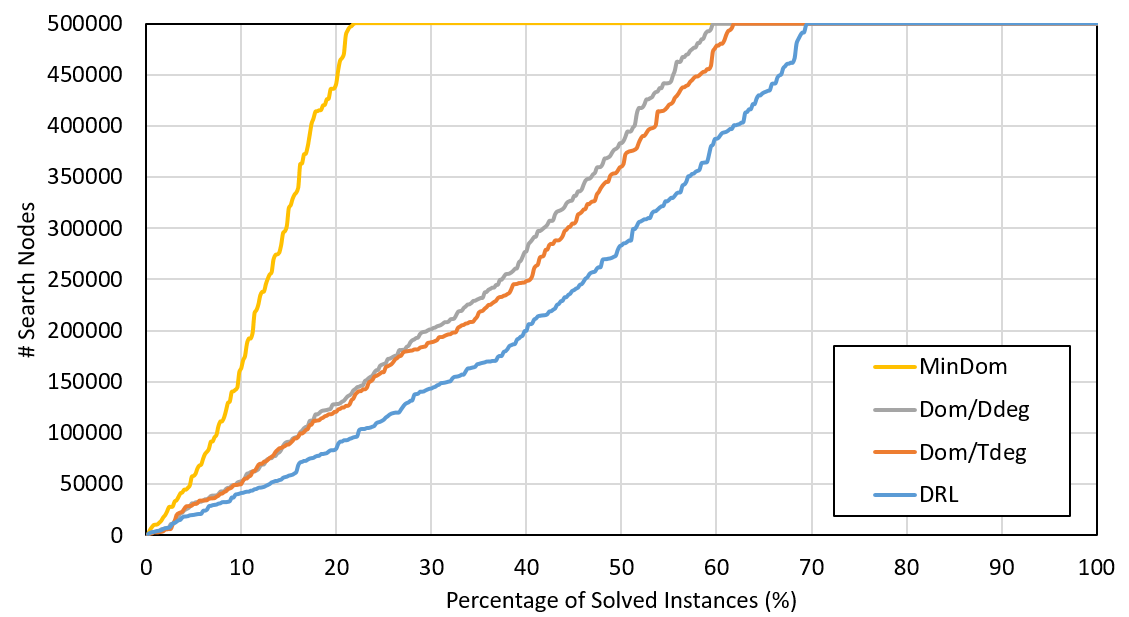}
		\caption{On distribution $\mathbb{D}_2(25)$}
		\label{fig:cactus2}
	\end{subfigure}
	\caption{Cactus plots of our method and baselines \textsf{Dom/Tdeg}, \textsf{Dom/Ddeg}, and \textsf{MinDom}.}
	\label{fig:cactus}
\end{figure}

\begin{figure}[!t]
	\centering
	\begin{subfigure}[b]{0.49\textwidth}
		\centering
		\includegraphics[width=\textwidth]{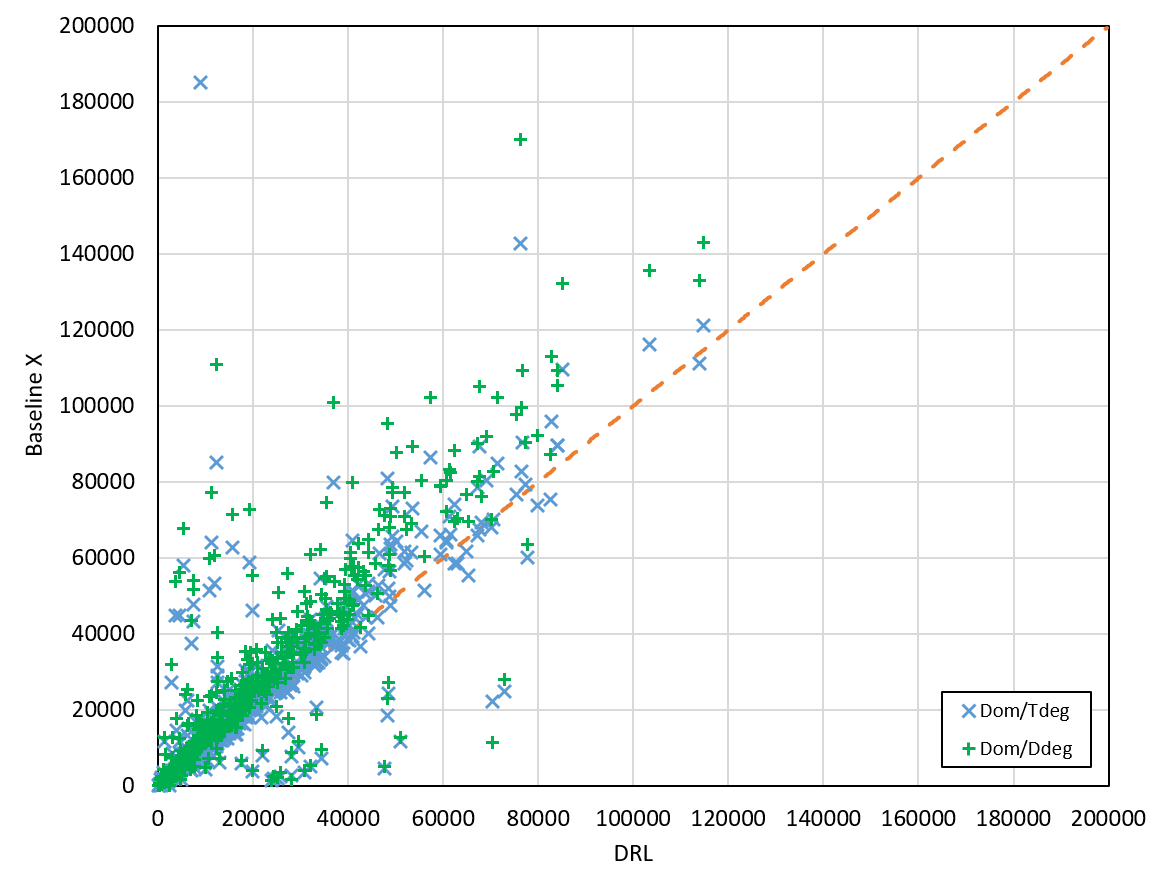}
		\caption{On distribution $\mathbb{D}_1(40)$}
		\label{fig:pairwise1}
	\end{subfigure}
	\begin{subfigure}[b]{0.49\textwidth}
		\centering
		\includegraphics[width=\textwidth]{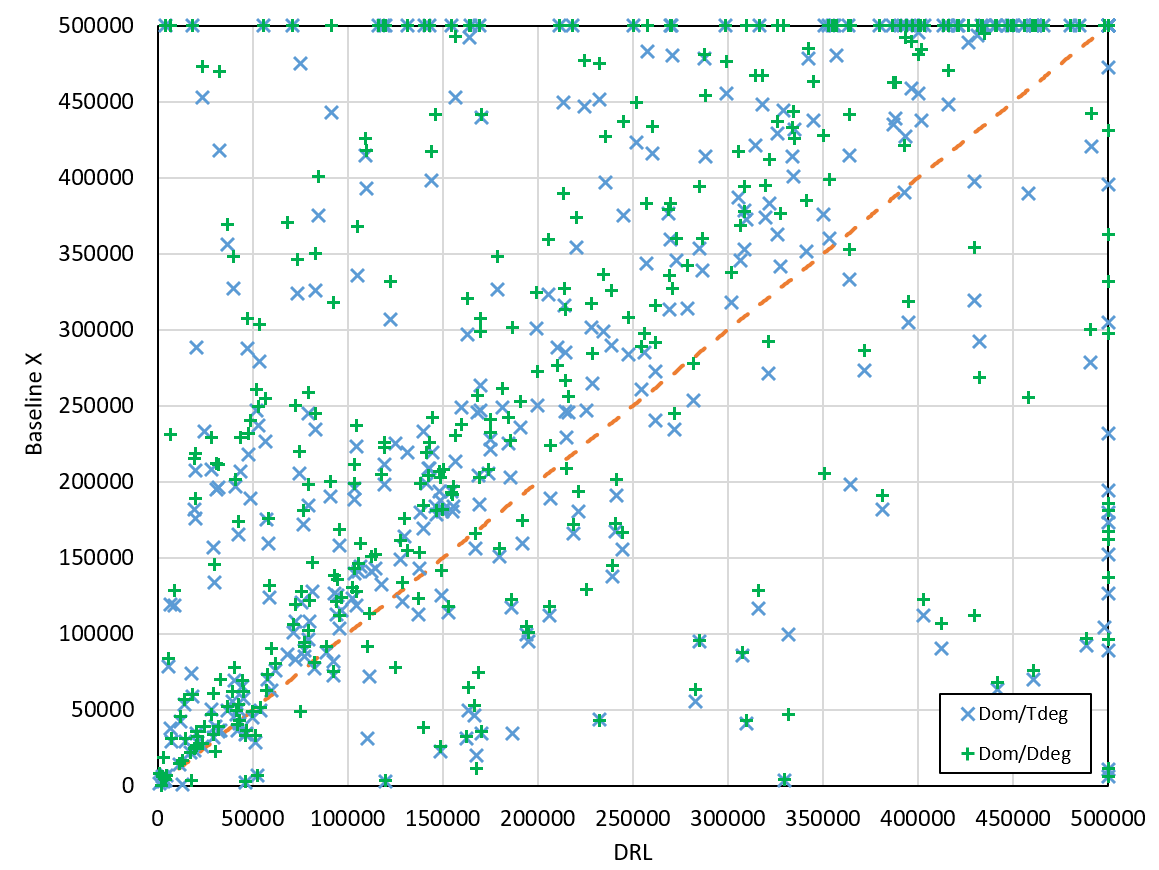}
		\caption{On distribution $\mathbb{D}_2(25)$}
		\label{fig:pairwise2}
	\end{subfigure}
	\caption{Pairwise comparison of Number of Search Nodes against Dom/Tdeg and Dom/Ddeg.}
	\label{fig:pairwise}
\end{figure}

To have a more fine-grained analysis and visualization of the aggregate results in Tables \ref{tb:gen_1} and \ref{tb:gen_2}, we show the cactus plots of our method and baselines on the hardest binary and 3-ary distribution $\mathbb{D}_1(40)$ and $\mathbb{D}_2(25)$ in Figure \ref{fig:cactus} (\textsf{Impact} is ignored due to the large number of cutoff instances). A cactus plot shows the increase of solved instances with the increase of solving resource bound (cutoff node limit here), and the lower and more to the right a curve is, the better the corresponding method. We can see that with the same cutoff node limit, our method almost consistently solves more instances than all baselines. Such improvement is more significant on the harder distribution $\mathbb{D}_2(25)$. To further validate the advantage of our method on the per-instance basis, we plot the pairwise comparisons against  \textsf{Dom/Tdeg} and \textsf{Dom/Ddeg} on $\mathbb{D}_1(40)$ and $\mathbb{D}_2(25)$ in Figure \ref{fig:pairwise}, where each dot is an instance and the coordinates are the corresponding search tree size of our method and the baseline. We can see that on both distributions, most dots are distributed above the diagonal (the dashed line), meaning that our method solves the corresponding instances with fewer nodes than the baselines.

To summarize, the above analysis shows that the policies trained by our approach have good generalization ability, and tend to perform better on harder instances with larger sizes and higher constraint arities, exhibiting even larger improvements than testing on instances of the same size as used in training. To support our observations and conclusion, we conduct statistical significant test to compare the performance of \textsf{DRL} and \textsf{Dom/Tdeg}.\footnote{Comparison to other baselines are also statistically significant, but the results are omitted here for brevity.} The $p$ values are listed in Table \ref{tb:pValue}, showing that all results are statistically significant ($p< 0.001$). 

\begin{table}[!t] \small
	\centering
	\caption{Statistical significant test results ($p$ value) for \textsf{DRL} and \textsf{Dom/Tdeg}}
	\label{tb:pValue}
	\begin{tabular}{ccc}
		\toprule
		$\mathbb{D}$ & \# Search Nodes    & \# Failures    \\ \midrule
		$\mathbb{D}_1(30)$ & 2.24$\times 10^{-12}$  & 2.28$\times 10^{-12}$ \\
		$\mathbb{D}_1(40)$ & 3.71$\times 10^{-20}$  & 3.66$\times 10^{-20}$  \\ \midrule
		$\mathbb{D}_2(20)$ & 3.64$\times 10^{-39}$ & 3.65$\times 10^{-39}$\\
		$\mathbb{D}_2(25)$ & 3.43$\times 10^{-13}$ & 3.43$\times 10^{-13}$ \\ \bottomrule
	\end{tabular}
\end{table}

\subsection{Impact of Different Neural Network Architecture} \label{sec:impact}
As mentioned previously, the network architecture and hyperparameters are tuned by performing grid search on small-sized instances, following the common practice in the literature. Here we show the impact of two hyperparameters, i.e. the number of GNN iteration $K$ and the number of MLP layers $L$, which we find to be crucial for the performance. We set $K\in \{1, 3, 5, 7\}$ and $L\in \{2, 3, 4, 5\}$, and perform training on the simplest distribution $\mathbb{D}_1(15)$ for each combination of $K$ and $L$. We also test the trained network on larger distribution $\mathbb{D}_1(20)$ to examine the generalization capability. As can be observed in Table \ref{tb:impact}, all combinations perform similarly on the training distribution. However, the generalization performance is significantly different. Specifically, with the increase of $K$, there is a notable trend that the search tree size on $\mathbb{D}_1(20)$ decreases, showing its great impact on generalization. This could be because with more iterations of aggregation and update, the variable and constraint embeddings could include richer information from neighbors located farther away on the constraint network. For each $K$, the increase of $L$ seems to be harmful, probably because of overfitting. The best performing combination is $K=5$ and $L=3$, which is also the configuration we used in the experiments.

\begin{table}[!t] \small
\centering
\caption{Impact of the number of GNN iteration $K$ and MLP layers $L$}
\label{tb:impact}
\begin{tabular}{@{}rrrrrr@{}}
\toprule
  \multirow{2}{*}{$K$} & \multirow{2}{*}{$L$} & \multicolumn{2}{c}{Training on  $\mathbb{D}_1(15)$}     & \multicolumn{2}{c}{Generalization on $\mathbb{D}_1(20)$}    \\ \cmidrule(lr){3-4} \cmidrule(lr){5-6}
 &  & \# Search Nodes & \# Failures    & \# Search Nodes & \# Failures    \\ \midrule
\multirow{4}{*}{1} & 2          & 25.63           & 10.35          & 329.7           & 162.13         \\
 & 3          & 25.63           & 10.35          & 718.81          & 356.57         \\
 & 4          & 26.51           & 10.80           & 190.45          & 92.15          \\
 & 5          & 25.59           & 10.28          & 212.71          & 103.21         \\ \midrule
\multirow{4}{*}{3} & 2          & 25.14           & 10.12          & 65.26           & 29.48          \\
 & 3          & 26.12           & 10.55          & 69.94           & 31.93          \\
 & 4          & 26.16           & 10.55          & 117.81          & 55.69          \\
 & 5          & 26.42           & 10.75          & 141.10           & 67.36          \\\midrule
\multirow{4}{*}{5} & 2          & 25.38           & 10.15          & 63.72           & 28.68          \\
 & 3          & 25.34           & 10.12          & \textBF{60.96}  & \textBF{27.34} \\
 & 4          & \textBF{25.08}  & 10.03          & 61.32           & 27.49          \\
 & 5          & 26.13           & 10.65          & 68.07           & 30.98          \\\midrule
\multirow{4}{*}{7} & 2          & 25.60            & 10.25          & 63.66           & 28.63          \\
 & 3          & 25.57           & 10.30           & 74.86           & 34.32          \\
 & 4          & 25.09           & \textBF{10.03} & 67.76           & 30.65          \\
 & 5          & 27.11           & 10.98          & 81.71           & 37.61          \\ \bottomrule
\end{tabular}
\end{table}

\begin{table}[!t] \scriptsize
	\centering
	\caption{Average execution time for all heuristics (all units are milliseconds unless stated)}
	\label{tb:time}
	\begin{tabular}{crrrrrrr}
		\toprule
		\multirow{2}{*}{$\mathbb{D}$} & \multicolumn{3}{c}{\textsf{DRL}} &
		\multirow{2}{*}{\textsf{Dom/Tdeg}} &
		\multirow{2}{*}{\textsf{Dom/Ddeg}} & \multirow{2}{*}{\textsf{MinDom}} & \multirow{2}{*}{\textsf{Impact}} \\ \cmidrule(lr){2-4}
		& Total  & Infer & No Infer &  &    &     &    \\ \midrule
		$\mathbb{D}_1(15)$                        & 23.47 & 98.32\% & 0.4  & 0.32  & \textbf{0.26}                & 0.40                    & 4.67                    \\
		$\mathbb{D}_1(25)$                        & 308.71 & 91.58\% & 26.01  & \textBF{23.04}   & 23.70                      & 48.62 & 2779.89 \\
		$\mathbb{D}_1(30)$                        & 1465.66 & 91.01\%  & 131.77   & 121.77  & \textBF{120.85}                    & 298.5                  & 14339.15                \\ 
		$\mathbb{D}_1(40)$                        & 36820.6 & 90.02\% & 3674.38   & \textBF{3881.52}  & 4132.99                   & 16451.9                & -                \\ \midrule
		$\mathbb{D}_2(10)$                        & 46.84 &97.14\% & 1.34    & 1.12  & \textBF{1.07}                      & 1.61                    & 5.81                     \\
		$\mathbb{D}_2(15)$                        & 788.92 & 93.54\% & 50.96   & 47.35  & \textbf{46.01}            & 95.36  & 1349.70                 \\
		$\mathbb{D}_2(20)$                        & 16883.9 & 91.97\% & 1356.42  & 1409.26  & \textBF{1371.18}    & 4378.82 & -                \\
		$\mathbb{D}_2(25)$              & 318995.48  & 89.17\%  & 34540.49 & 35581.39 & \textBF{33515.28}                  & 43539.38                  & -                \\ \bottomrule
	\end{tabular}
\end{table}

\subsection{Execution Time Analysis} \label{sec:timeAnalysis}

We finally analyze the execution times of all the heuristics, which are listed in Table \ref{tb:time} (some results of \textsf{Impact} are ignored because a majority of instances are cut off). We can observe that the total execution times of \textsf{DRL} (shown in the ``Total" column) are much longer than those of the baselines. As shown in the ``Infer" column, almost over 90\% of the execution time is spent in GNN inference. In contrast, the times for variable ordering decisions of the baseline heuristics are negligible since the related computation is very simple and fast. With the increase of problem size, the percentage of inference time decreases, showing that more portion of time is spent in ``actually" solving the instances (mostly constraint propagation). To measure these efforts, we calculate the execution times of \textsf{DRL} without inference in the ``No Infer" column. We can observe that the ``No Infer" time of \textsf{DRL} is generally much short than the run time of \textsf{MinDom} and \textsf{Impact}, the two embedded heuristics in OR-tools, and comparable to those of \textsf{Dom/Ddeg} and \textsf{Dom/Tdeg}.

\begin{table} \small
	\centering
	\caption{Performance of inferring on the topmost $\mathcal{K}$ nodes on $\mathbb{D}_1(40)$}
	\label{tb:tradeoff_2}
	\begin{threeparttable}
	\begin{tabular}{crrrrr}
		\toprule
		\multirow{2}{*}{Heuristic} & \multicolumn{2}{c}{\# Search Nodes} & \multirow{2}{*}{\# Cutoff} &	\multicolumn{2}{c}{Time (milliseconds)}
		\\ 
		\cmidrule(lr){2-3}
		\cmidrule(lr){5-6}
		& Average   & Reduction &  & Average  & Ratio ($\times$)
		\\ \midrule
		\textsf{Dom/Tdeg}                        & 26861.89     & - & - & 3881.52  & -  \\ 
		\textsf{DRL} & 23684.04 & 11.83\% & - & 36820.63 & 9.486 \\
		\midrule
		\textsf{DRL ($\mathcal{K}$=15)}                   & \textBF{23327.71} & 13.16\%         &  -   & 14245.39 & 3.670 \\
		\textsf{DRL ($\mathcal{K}$=10)}                   & 23967.82  & 10.77\%  & - & 5840.75 &
		1.505 \\
		\textsf{DRL ($\mathcal{K}$=5)}                   & 24983.85 & 6.99\%  & -  & 3793.13 & 0.977        \\
		\textsf{DRL ($\mathcal{K}$=3)}                   & 25402.32 & 5.43\%  & -  & \textBF{3698.68} & 0.953          \\
		\bottomrule
	\end{tabular}
	\begin{tablenotes}
	\footnotesize
	\item Note: Reduction and Ratio values are computed using instances solved by both \textsf{DRL} and \textsf{Dom/Tdeg}.
	\end{tablenotes}
	\end{threeparttable}
\end{table}

\begin{table} \small
	\centering
	\caption{Performance of inferring on the topmost $\mathcal{K}$ nodes on $\mathbb{D}_2(25)$}
	\label{tb:tradeoff_3}
	\begin{threeparttable}
	\begin{tabular}{crrrrr}
		\toprule
		\multirow{2}{*}{Heuristic} & \multicolumn{2}{c}{\# Search Nodes} & \multirow{2}{*}{\# Cutoff} &	\multicolumn{2}{c}{Time (milliseconds)}
		\\ 
		\cmidrule(lr){2-3}
		\cmidrule(lr){5-6}
		& Average   & Reduction &  & Average  & Ratio ($\times$)  \\ \midrule
		\textsf{Dom/Tdeg}                        & 322161.17     & -    & 191 & 35581.39 & -  \\ 
		\textsf{DRL} & \textBF{287406.14} & 18.74\% & \textBF{153} & 329637.97 & 8.174 \\
		\midrule
		\textsf{DRL ($\mathcal{K}$=15)}                   & 294922.66 & 14.81\% & 159    & 55643.28 & 1.651 \\
		\textsf{DRL ($\mathcal{K}$=10)}                   & 302834.59  & 6.00\%         & 169    & 35317.73 & 0.962 \\
		\textsf{DRL ($\mathcal{K}$=5)}                   & 309851.98 & 3.82\% & 179    & \textBF{34302.71} & 0.921  \\
		\textsf{DRL ($\mathcal{K}$=3)}                   & 312273.33 & 3.07\%  & 181    & 34463.92 & 0.933          \\
		\bottomrule
	\end{tabular}
	\begin{tablenotes}
		\footnotesize
		\item Note: Reduction and Ratio values are computed using instances solved by both \textsf{DRL} and \textsf{Dom/Tdeg}.
	\end{tablenotes}
	\end{threeparttable}
\end{table}

Note that it is common in tree search algorithms that high-quality heuristics leading to smaller search tree consume  much longer computational time, such as the well-known and effective heuristic Strong Branching for MILP \citep{applegate1995finding}. Nevertheless, minimizing search tree size is a valid and well-recognized objective for both traditional and learning-based methods \citep{liberatore2000complexity,hooker1995testing,jo2021param}. Our objective in this paper is not to compete with traditional heuristics in execution time, but to show that high-quality search policies for CSP can be automatically learned in a data-driven way, by a carefully designed deep RL method. We believe the efficiency of our method could be significantly improved with more advanced techniques such as parameter pruning and model compression, but is out of the scope of this paper and is a direction for future research.

Here we show a simple strategy that maintains the advantage of our method in producing smaller search trees, but is comparable and even faster than traditional heuristics in execution time.  Specifically, based on the intuition that earlier decisions in the search process are more critical, we perform DNN inference on nodes in the topmost $\mathcal{K}$ layers of the search tree, and use traditional heuristics for the other nodes. We perform experiments on the hardest distributions $\mathbb{D}_1(40)$ and $\mathbb{D}_2(25)$ for binary and 3-ary problems. We choose $\mathcal{K}\in \{3,5,10,15\}$, and use the best-performing traiditional heuristic \textsf{Dom/Tdeg} for the remaining nodes. Results are summarized in Tables \ref{tb:tradeoff_2} and \ref{tb:tradeoff_3}, where the ``Ratio" column is the ratio of execution time of \textsf{DRL} to that of \textsf{Dom/Tdeg}, computed on the commonly solved instances. We can observe that this strategy significantly reduces the execution time. With the decrease of $\mathcal{K}$, the speed-up is more prominent while the reduction in search tree size becomes smaller, which is an expected trade-off effect. On $\mathcal{D}_1(40)$, \textsf{DRL} starts to outperform \textsf{Dom/Tdeg} in execution time when $\mathcal{K}=5$, and is 4.3\% faster than \textsf{Dom/Tdeg} when $\mathcal{K}=3$ while still maintaining 5.43\% reduction in the search tree size. On the harder distribution $\mathcal{D}_2(25)$, \textsf{DRL} starts to be faster than \textsf{Dom/Tdeg} when $\mathcal{K}=10$. The fastest version is \textsf{DRL ($\mathcal{K}=5$)}, which is 7.9\% faster than \textsf{Dom/Tdeg} on the solved instances with 3.82\% reduction in the search tree size.

\section{Discussions and Limitations} 
\label{sec:discussion}

Experimental results in Section \ref{sec:experiments} show that in most cases, the variable ordering heuristics learned by our DRL method are able to outperform traditional hand-crafted ones in minimizing the search tree size. This advantage could come from the following important facts. First, in our method, the ordering decisions are made based on a learned Q function which is explicitly optimized towards minimization of the search tree, guided by the surrogate objective defined in Equation (\ref{eq:qhat}). In contrast, the hand-crafted heuristics are designed based on human experience, which have no direct relation with the search effort. Second, traditional heuristics make decisions based on only simple features (e.g. domain and dynamic degree of each variable). In contrast, for our method, the features extracted by our GNN based scheme is more informative and comprehensive, which provides more accurate solving status so that better decisions could be made. Finally, the size-agnostic property of GNN makes it possible to reuse the knowledge learned on small instances to larger ones. Also, the message passing mechanism of GNN captures the local patterns of the underlying state graph, which could also apply to larger instances.

While showing promising results, our method also has several limitations that need to be improved in the future for practical usage. First, currently we only consider table constraints. While they can theoretically encode any constraints, in practise the constraints could be defined more compactly in intension. One solution to this issue is to extend our method with type-aware constraint representation which could allow heterogeneous edge features, so as to model more constraint types. Second, as we have analyzed in Section \ref{sec:timeAnalysis}, the inference time of GNN is a bottleneck of our method. Similar situations have also been noticed in other recent works, e.g. \cite{cappart2021combining,chalumeau2021seapearl}. While we have shown a simple strategy in Section \ref{sec:timeAnalysis} (inferring GNN on the topmost search nodes) could partially resolve this issue, more principled and rigorous design is needed to achieve desired trade-off between model accuracy and efficiency, such as the hybrid scheme in \citep{gupta2020hybrid}.

\section{Conclusions and Future Work} \label{sec:conclusions}
In this paper, we make a preliminary attempt on tackling CSP using deep reinforcement learning. Specifically, we study the problem of how to use DRL to automatically discover a variable ordering heuristic for a given class of CSP instances. We propose a reinforcement learning formulation for this task, which allows the decision making agent to learn from its own solving experiences without the need of supervision. 
Extensive experiments show that our RL agent can discover variable ordering policies that are better than traditional hand-craft heuristics, in terms of minimizing the search tree size. The learned policies is size agnostic, and can generalize to instances that are larger than those used for training. Moreover, the improvement over traditional heuristics tends to be more significant on larger and harder problems. The framework we designed could also be applied to learn other control policies in backtracking search algorithms, such as value ordering, propagator selection, and determining propagation level. In the future, we plan to improve and extend our method to enhance practical CSP solving in real industrial applications, such as resource scheduling, industrial robot control, and logistics planning.

\section*{Acknowledgement}
This study is supported under the RIE2020 Industry Alignment Fund – Industry Collaboration Projects (IAF-ICP) Funding Initiative, as well as cash and in-kind contribution from Singapore Telecommunications Limited (Singtel), through Singtel Cognitive and Artificial Intelligence Lab for Enterprises (SCALE@NTU). This study is also supported by the National Natural Science Foundation of China under Grant 62102228 and 61803104, and in part by Shandong Provincial Natural Science Foundation under Grant ZR2021QF063, and in part by the A*STAR Cyber-Physical Production System (CPPS) – Towards Contextual and Intelligent Response Research Program, under the RIE2020 IAF-PP Grant A19C1a0018.

\bibliography{learning_heuristic}

\begin{thebibliography}{55}
\expandafter\ifx\csname natexlab\endcsname\relax\def\natexlab#1{#1}\fi
\providecommand{\url}[1]{\texttt{#1}}
\providecommand{\href}[2]{#2}
\providecommand{\path}[1]{#1}
\providecommand{\DOIprefix}{doi:}
\providecommand{\ArXivprefix}{arXiv:}
\providecommand{\URLprefix}{URL: }
\providecommand{\Pubmedprefix}{pmid:}
\providecommand{\doi}[1]{\href{http://dx.doi.org/#1}{\path{#1}}}
\providecommand{\Pubmed}[1]{\href{pmid:#1}{\path{#1}}}
\providecommand{\bibinfo}[2]{#2}
\ifx\xfnm\relax \def\xfnm[#1]{\unskip,\space#1}\fi
\bibitem[{Altan et~al.(2021)Altan, Karasu \& Zio}]{altan2021new}
\bibinfo{author}{Altan, A.}, \bibinfo{author}{Karasu, S.}, \&
  \bibinfo{author}{Zio, E.} (\bibinfo{year}{2021}).
\newblock \bibinfo{title}{A new hybrid model for wind speed forecasting
  combining long short-term memory neural network, decomposition methods and
  grey wolf optimizer}.
\newblock {\it \bibinfo{journal}{Applied Soft Computing}\/},  {\it
  \bibinfo{volume}{100}\/}, \bibinfo{pages}{106996}.
\bibitem[{Amizadeh et~al.(2019)Amizadeh, Matusevych \&
  Weimer}]{amizadeh2018learning}
\bibinfo{author}{Amizadeh, S.}, \bibinfo{author}{Matusevych, S.}, \&
  \bibinfo{author}{Weimer, M.} (\bibinfo{year}{2019}).
\newblock \bibinfo{title}{Learning to solve circuit-sat: An unsupervised
  differentiable approach}.
\newblock In {\it \bibinfo{booktitle}{International Conference on Learning
  Representations}\/}.
\bibitem[{Applegate et~al.(1995)Applegate, Bixby, Chvatal \&
  Cook}]{applegate1995finding}
\bibinfo{author}{Applegate, D.}, \bibinfo{author}{Bixby, R.},
  \bibinfo{author}{Chvatal, V.}, \& \bibinfo{author}{Cook, B.}
  (\bibinfo{year}{1995}).
\newblock \bibinfo{title}{Finding cuts in the tsp (a preliminary report)}.
\bibitem[{Balafrej et~al.(2015)Balafrej, Bessiere \&
  Paparrizou}]{balafrej2015multi}
\bibinfo{author}{Balafrej, A.}, \bibinfo{author}{Bessiere, C.}, \&
  \bibinfo{author}{Paparrizou, A.} (\bibinfo{year}{2015}).
\newblock \bibinfo{title}{Multi-armed bandits for adaptive constraint
  propagation}.
\newblock In {\it \bibinfo{booktitle}{International Joint Conference on
  Artificial Intelligence}\/} (pp. \bibinfo{pages}{290--296}).
\bibitem[{Beck et~al.(2004)Beck, Prosser \& Wallace}]{beck2004trying}
\bibinfo{author}{Beck, J.~C.}, \bibinfo{author}{Prosser, P.}, \&
  \bibinfo{author}{Wallace, R.~J.} (\bibinfo{year}{2004}).
\newblock \bibinfo{title}{Trying again to fail-first}.
\newblock In {\it \bibinfo{booktitle}{International Workshop on Constraint
  Solving and Constraint Logic Programming}\/} (pp. \bibinfo{pages}{41--55}).
\newblock \bibinfo{organization}{Springer}.
\bibitem[{Behrens et~al.(2019)Behrens, Lange \&
  Mansouri}]{behrens2019constraint}
\bibinfo{author}{Behrens, J.~K.}, \bibinfo{author}{Lange, R.}, \&
  \bibinfo{author}{Mansouri, M.} (\bibinfo{year}{2019}).
\newblock \bibinfo{title}{A constraint programming approach to simultaneous
  task allocation and motion scheduling for industrial dual-arm manipulation
  tasks}.
\newblock In {\it \bibinfo{booktitle}{2019 International Conference on Robotics
  and Automation (ICRA)}\/} (pp. \bibinfo{pages}{8705--8711}).
\newblock \bibinfo{organization}{IEEE}.
\bibitem[{Bengio et~al.(2020)Bengio, Lodi \& Prouvost}]{bengio2020machine}
\bibinfo{author}{Bengio, Y.}, \bibinfo{author}{Lodi, A.}, \&
  \bibinfo{author}{Prouvost, A.} (\bibinfo{year}{2020}).
\newblock \bibinfo{title}{Machine learning for combinatorial optimization: a
  methodological tour d’horizon}.
\newblock {\it \bibinfo{journal}{European Journal of Operational Research}\/},
  {\it \bibinfo{volume}{290}\/}, \bibinfo{pages}{405--421}.
\bibitem[{Bessiere \& R{\'e}gin(1996)}]{bessiere1996mac}
\bibinfo{author}{Bessiere, C.}, \& \bibinfo{author}{R{\'e}gin, J.-C.}
  (\bibinfo{year}{1996}).
\newblock \bibinfo{title}{Mac and combined heuristics: Two reasons to forsake
  fc (and cbj?) on hard problems}.
\newblock In {\it \bibinfo{booktitle}{International Conference on Principles
  and Practice of Constraint Programming}\/} (pp. \bibinfo{pages}{61--75}).
\newblock \bibinfo{organization}{Springer}.
\bibitem[{Cappart et~al.(2021)Cappart, Moisan, Rousseau, Pr{\'e}mont-Schwarz \&
  Cire}]{cappart2021combining}
\bibinfo{author}{Cappart, Q.}, \bibinfo{author}{Moisan, T.},
  \bibinfo{author}{Rousseau, L.-M.}, \bibinfo{author}{Pr{\'e}mont-Schwarz, I.},
  \& \bibinfo{author}{Cire, A.~A.} (\bibinfo{year}{2021}).
\newblock \bibinfo{title}{Combining reinforcement learning and constraint
  programming for combinatorial optimization}.
\newblock In {\it \bibinfo{booktitle}{Proceedings of the AAAI Conference on
  Artificial Intelligence}\/} (pp. \bibinfo{pages}{3677--3687}).
\newblock volume~\bibinfo{volume}{35}.
\bibitem[{Chalumeau et~al.(2021)Chalumeau, Coulon, Cappart \&
  Rousseau}]{chalumeau2021seapearl}
\bibinfo{author}{Chalumeau, F.}, \bibinfo{author}{Coulon, I.},
  \bibinfo{author}{Cappart, Q.}, \& \bibinfo{author}{Rousseau, L.-M.}
  (\bibinfo{year}{2021}).
\newblock \bibinfo{title}{Seapearl: A constraint programming solver guided by
  reinforcement learning}.
\newblock In {\it \bibinfo{booktitle}{International Conference on Integration
  of Constraint Programming, Artificial Intelligence, and Operations
  Research}\/} (pp. \bibinfo{pages}{392--409}).
\newblock \bibinfo{organization}{Springer}.
\bibitem[{Correia \& Barahona(2008)}]{correia2008efficiency}
\bibinfo{author}{Correia, M.}, \& \bibinfo{author}{Barahona, P.}
  (\bibinfo{year}{2008}).
\newblock \bibinfo{title}{On the efficiency of impact based heuristics}.
\newblock In {\it \bibinfo{booktitle}{International Conference on Principles
  and Practice of Constraint Programming}\/} (pp. \bibinfo{pages}{608--612}).
\newblock \bibinfo{organization}{Springer}.
\bibitem[{Demeulenaere et~al.(2016)Demeulenaere, Hartert, Lecoutre, Perez,
  Perron, R{\'e}gin \& Schaus}]{demeulenaere2016compact}
\bibinfo{author}{Demeulenaere, J.}, \bibinfo{author}{Hartert, R.},
  \bibinfo{author}{Lecoutre, C.}, \bibinfo{author}{Perez, G.},
  \bibinfo{author}{Perron, L.}, \bibinfo{author}{R{\'e}gin, J.-C.}, \&
  \bibinfo{author}{Schaus, P.} (\bibinfo{year}{2016}).
\newblock \bibinfo{title}{Compact-table: efficiently filtering table
  constraints with reversible sparse bit-sets}.
\newblock In {\it \bibinfo{booktitle}{International Conference on Principles
  and Practice of Constraint Programming}\/} (pp. \bibinfo{pages}{207--223}).
\newblock \bibinfo{organization}{Springer}.
\bibitem[{Ding et~al.(2020)Ding, Zhang, Shen, Li, Wang, Xu \&
  Song}]{ding2020accelerating}
\bibinfo{author}{Ding, J.-Y.}, \bibinfo{author}{Zhang, C.},
  \bibinfo{author}{Shen, L.}, \bibinfo{author}{Li, S.}, \bibinfo{author}{Wang,
  B.}, \bibinfo{author}{Xu, Y.}, \& \bibinfo{author}{Song, L.}
  (\bibinfo{year}{2020}).
\newblock \bibinfo{title}{Accelerating primal solution findings for mixed
  integer programs based on solution prediction.}
\newblock In {\it \bibinfo{booktitle}{Proceedings of the Thirty-Fourth AAAI
  Conference on Artificial Intelligence}\/} (pp. \bibinfo{pages}{1452--1459}).
\bibitem[{Epstein \& Petrovic(2007)}]{epstein2007learning}
\bibinfo{author}{Epstein, S.}, \& \bibinfo{author}{Petrovic, S.}
  (\bibinfo{year}{2007}).
\newblock \bibinfo{title}{Learning to solve constraint problems}.
\newblock In {\it \bibinfo{booktitle}{ICAPS-07 Workshop on Planning and
  Learning}\/}.
\bibitem[{Galassi et~al.(2018)Galassi, Lombardi, Mello \&
  Milano}]{galassi2018model}
\bibinfo{author}{Galassi, A.}, \bibinfo{author}{Lombardi, M.},
  \bibinfo{author}{Mello, P.}, \& \bibinfo{author}{Milano, M.}
  (\bibinfo{year}{2018}).
\newblock \bibinfo{title}{Model agnostic solution of csps via deep learning: A
  preliminary study}.
\newblock In {\it \bibinfo{booktitle}{International Conference on the
  Integration of Constraint Programming, Artificial Intelligence, and
  Operations Research}\/} (pp. \bibinfo{pages}{254--262}).
\newblock \bibinfo{organization}{Springer}.
\bibitem[{Gasse et~al.(2019)Gasse, Ch{\'e}telat, Ferroni, Charlin \&
  Lodi}]{gasse2019exact}
\bibinfo{author}{Gasse, M.}, \bibinfo{author}{Ch{\'e}telat, D.},
  \bibinfo{author}{Ferroni, N.}, \bibinfo{author}{Charlin, L.}, \&
  \bibinfo{author}{Lodi, A.} (\bibinfo{year}{2019}).
\newblock \bibinfo{title}{Exact combinatorial optimization with graph
  convolutional neural networks}.
\newblock In {\it \bibinfo{booktitle}{Advances in Neural Information Processing
  Systems}\/} (pp. \bibinfo{pages}{15580--15592}).
\bibitem[{Gent et~al.(1996)Gent, MacIntyre, Presser, Smith \&
  Walsh}]{gent1996empirical}
\bibinfo{author}{Gent, I.~P.}, \bibinfo{author}{MacIntyre, E.},
  \bibinfo{author}{Presser, P.}, \bibinfo{author}{Smith, B.~M.}, \&
  \bibinfo{author}{Walsh, T.} (\bibinfo{year}{1996}).
\newblock \bibinfo{title}{An empirical study of dynamic variable ordering
  heuristics for the constraint satisfaction problem}.
\newblock In {\it \bibinfo{booktitle}{International Conference on Principles
  and Practice of Constraint Programming}\/} (pp. \bibinfo{pages}{179--193}).
\newblock \bibinfo{organization}{Springer}.
\bibitem[{Gupta et~al.(2020)Gupta, Gasse, Khalil, Mudigonda, Lodi \&
  Bengio}]{gupta2020hybrid}
\bibinfo{author}{Gupta, P.}, \bibinfo{author}{Gasse, M.},
  \bibinfo{author}{Khalil, E.}, \bibinfo{author}{Mudigonda, P.},
  \bibinfo{author}{Lodi, A.}, \& \bibinfo{author}{Bengio, Y.}
  (\bibinfo{year}{2020}).
\newblock \bibinfo{title}{Hybrid models for learning to branch}.
\newblock {\it \bibinfo{journal}{Advances in Neural Information Processing
  Systems}\/},  {\it \bibinfo{volume}{33}\/}, \bibinfo{pages}{18087--18097}.
\bibitem[{Haralick \& Elliott(1980)}]{haralick1980increasing}
\bibinfo{author}{Haralick, R.~M.}, \& \bibinfo{author}{Elliott, G.~L.}
  (\bibinfo{year}{1980}).
\newblock \bibinfo{title}{Increasing tree search efficiency for constraint
  satisfaction problems}.
\newblock {\it \bibinfo{journal}{Artificial intelligence}\/},  {\it
  \bibinfo{volume}{14}\/}, \bibinfo{pages}{263--313}.
\bibitem[{Hasselt et~al.(2016)Hasselt, Guez \& Silver}]{van2016deep}
\bibinfo{author}{Hasselt, H.~v.}, \bibinfo{author}{Guez, A.}, \&
  \bibinfo{author}{Silver, D.} (\bibinfo{year}{2016}).
\newblock \bibinfo{title}{Deep reinforcement learning with double q-learning}.
\newblock In {\it \bibinfo{booktitle}{Proceedings of the Thirtieth AAAI
  Conference on Artificial Intelligence}\/} (pp. \bibinfo{pages}{2094--2100}).
\bibitem[{He et~al.(2014)He, Daume~III \& Eisner}]{he2014learning}
\bibinfo{author}{He, H.}, \bibinfo{author}{Daume~III, H.}, \&
  \bibinfo{author}{Eisner, J.~M.} (\bibinfo{year}{2014}).
\newblock \bibinfo{title}{Learning to search in branch and bound algorithms}.
\newblock In {\it \bibinfo{booktitle}{Advances in neural information processing
  systems}\/} (pp. \bibinfo{pages}{3293--3301}).
\bibitem[{Hooker(1995)}]{hooker1995testing}
\bibinfo{author}{Hooker, J.~N.} (\bibinfo{year}{1995}).
\newblock \bibinfo{title}{Testing heuristics: We have it all wrong}.
\newblock {\it \bibinfo{journal}{Journal of heuristics}\/},  {\it
  \bibinfo{volume}{1}\/}, \bibinfo{pages}{33--42}.
\bibitem[{Kasprzak et~al.(2014)Kasprzak, Szynkiewicz, Zlatanov \&
  Zieli{\'n}ska}]{kasprzak2014hierarchical}
\bibinfo{author}{Kasprzak, W.}, \bibinfo{author}{Szynkiewicz, W.},
  \bibinfo{author}{Zlatanov, D.}, \& \bibinfo{author}{Zieli{\'n}ska, T.}
  (\bibinfo{year}{2014}).
\newblock \bibinfo{title}{A hierarchical csp search for path planning of
  cooperating self-reconfigurable mobile fixtures}.
\newblock {\it \bibinfo{journal}{Engineering Applications of Artificial
  Intelligence}\/},  {\it \bibinfo{volume}{34}\/}, \bibinfo{pages}{85--98}.
\bibitem[{Khalil et~al.(2017{\natexlab{a}})Khalil, Dai, Zhang, Dilkina \&
  Song}]{khalil2017learning}
\bibinfo{author}{Khalil, E.}, \bibinfo{author}{Dai, H.},
  \bibinfo{author}{Zhang, Y.}, \bibinfo{author}{Dilkina, B.}, \&
  \bibinfo{author}{Song, L.} (\bibinfo{year}{2017}{\natexlab{a}}).
\newblock \bibinfo{title}{Learning combinatorial optimization algorithms over
  graphs}.
\newblock In {\it \bibinfo{booktitle}{Advances in Neural Information Processing
  Systems}\/} (pp. \bibinfo{pages}{6348--6358}).
\bibitem[{Khalil et~al.(2017{\natexlab{b}})Khalil, Dilkina, Nemhauser, Ahmed \&
  Shao}]{khalil2017learningRun}
\bibinfo{author}{Khalil, E.~B.}, \bibinfo{author}{Dilkina, B.},
  \bibinfo{author}{Nemhauser, G.~L.}, \bibinfo{author}{Ahmed, S.}, \&
  \bibinfo{author}{Shao, Y.} (\bibinfo{year}{2017}{\natexlab{b}}).
\newblock \bibinfo{title}{Learning to run heuristics in tree search}.
\newblock In {\it \bibinfo{booktitle}{International Joint Conference on
  Artificial Intelligence}\/} (pp. \bibinfo{pages}{659--666}).
\bibitem[{Khalil et~al.(2016)Khalil, Le~Bodic, Song, Nemhauser \&
  Dilkina}]{khalil2016learning}
\bibinfo{author}{Khalil, E.~B.}, \bibinfo{author}{Le~Bodic, P.},
  \bibinfo{author}{Song, L.}, \bibinfo{author}{Nemhauser, G.}, \&
  \bibinfo{author}{Dilkina, B.} (\bibinfo{year}{2016}).
\newblock \bibinfo{title}{Learning to branch in mixed integer programming}.
\newblock In {\it \bibinfo{booktitle}{Thirtieth AAAI Conference on Artificial
  Intelligence}\/} (pp. \bibinfo{pages}{724--731}).
\bibitem[{Kool et~al.(2019)Kool, van Hoof \& Welling}]{kool2018attention}
\bibinfo{author}{Kool, W.}, \bibinfo{author}{van Hoof, H.}, \&
  \bibinfo{author}{Welling, M.} (\bibinfo{year}{2019}).
\newblock \bibinfo{title}{Attention, learn to solve routing problems!}
\newblock In {\it \bibinfo{booktitle}{International Conference on Learning
  Representations}\/}.
\bibitem[{Lagoudakis \& Littman(2000)}]{lagoudakis2000algorithm}
\bibinfo{author}{Lagoudakis, M.~G.}, \& \bibinfo{author}{Littman, M.~L.}
  (\bibinfo{year}{2000}).
\newblock \bibinfo{title}{Algorithm selection using reinforcement learning}.
\newblock In {\it \bibinfo{booktitle}{International Conference on Machine
  Learning}\/} (pp. \bibinfo{pages}{511--518}).
\bibitem[{Lagoudakis \& Littman(2001)}]{lagoudakis2001learning}
\bibinfo{author}{Lagoudakis, M.~G.}, \& \bibinfo{author}{Littman, M.~L.}
  (\bibinfo{year}{2001}).
\newblock \bibinfo{title}{Learning to select branching rules in the dpll
  procedure for satisfiability}.
\newblock {\it \bibinfo{journal}{Electronic Notes in Discrete Mathematics}\/},
  {\it \bibinfo{volume}{9}\/}, \bibinfo{pages}{344--359}.
\bibitem[{Legat \& Vogel-Heuser(2017)}]{legat2017configurable}
\bibinfo{author}{Legat, C.}, \& \bibinfo{author}{Vogel-Heuser, B.}
  (\bibinfo{year}{2017}).
\newblock \bibinfo{title}{A configurable partial-order planning approach for
  field level operation strategies of plc-based industry 4.0 automated
  manufacturing systems}.
\newblock {\it \bibinfo{journal}{Engineering Applications of Artificial
  Intelligence}\/},  {\it \bibinfo{volume}{66}\/}, \bibinfo{pages}{128--144}.
\bibitem[{Li et~al.(2016)Li, Liang, Zhang, Guo, Xu \& Li}]{li2016improving}
\bibinfo{author}{Li, H.}, \bibinfo{author}{Liang, Y.}, \bibinfo{author}{Zhang,
  N.}, \bibinfo{author}{Guo, J.}, \bibinfo{author}{Xu, D.}, \&
  \bibinfo{author}{Li, Z.} (\bibinfo{year}{2016}).
\newblock \bibinfo{title}{Improving degree-based variable ordering heuristics
  for solving constraint satisfaction problems}.
\newblock {\it \bibinfo{journal}{Journal of Heuristics}\/},  {\it
  \bibinfo{volume}{22}\/}, \bibinfo{pages}{125--145}.
\bibitem[{Li et~al.(2018)Li, Chen \& Koltun}]{li2018combinatorial}
\bibinfo{author}{Li, Z.}, \bibinfo{author}{Chen, Q.}, \&
  \bibinfo{author}{Koltun, V.} (\bibinfo{year}{2018}).
\newblock \bibinfo{title}{Combinatorial optimization with graph convolutional
  networks and guided tree search}.
\newblock In {\it \bibinfo{booktitle}{Advances in Neural Information Processing
  Systems}\/} (pp. \bibinfo{pages}{539--548}).
\bibitem[{Liberatore(2000)}]{liberatore2000complexity}
\bibinfo{author}{Liberatore, P.} (\bibinfo{year}{2000}).
\newblock \bibinfo{title}{On the complexity of choosing the branching literal
  in dpll}.
\newblock {\it \bibinfo{journal}{Artificial intelligence}\/},  {\it
  \bibinfo{volume}{116}\/}, \bibinfo{pages}{315--326}.
\bibitem[{Ma et~al.(2020)Ma, Liu, Cao, Song, Zhang \& Zeng}]{ma2020cost}
\bibinfo{author}{Ma, C.}, \bibinfo{author}{Liu, Z.}, \bibinfo{author}{Cao, Z.},
  \bibinfo{author}{Song, W.}, \bibinfo{author}{Zhang, J.}, \&
  \bibinfo{author}{Zeng, W.} (\bibinfo{year}{2020}).
\newblock \bibinfo{title}{Cost-sensitive deep forest for price prediction}.
\newblock {\it \bibinfo{journal}{Pattern Recognition}\/},  {\it
  \bibinfo{volume}{107}\/}, \bibinfo{pages}{107499}.
\bibitem[{Mackworth \& Freuder(1993)}]{mackworth1993complexity}
\bibinfo{author}{Mackworth, A.~K.}, \& \bibinfo{author}{Freuder, E.~C.}
  (\bibinfo{year}{1993}).
\newblock \bibinfo{title}{The complexity of constraint satisfaction revisited}.
\newblock {\it \bibinfo{journal}{Artificial Intelligence}\/},  {\it
  \bibinfo{volume}{59}\/}, \bibinfo{pages}{57--62}.
\bibitem[{Mao et~al.(2019)Mao, Schwarzkopf, Venkatakrishnan, Meng \&
  Alizadeh}]{mao2019learning}
\bibinfo{author}{Mao, H.}, \bibinfo{author}{Schwarzkopf, M.},
  \bibinfo{author}{Venkatakrishnan, S.~B.}, \bibinfo{author}{Meng, Z.}, \&
  \bibinfo{author}{Alizadeh, M.} (\bibinfo{year}{2019}).
\newblock \bibinfo{title}{Learning scheduling algorithms for data processing
  clusters}.
\newblock In {\it \bibinfo{booktitle}{Proceedings of the ACM Special Interest
  Group on Data Communication}\/} (pp. \bibinfo{pages}{270--288}).
\bibitem[{Pardo et~al.(2018)Pardo, Tavakoli, Levdik \&
  Kormushev}]{pardo2018time}
\bibinfo{author}{Pardo, F.}, \bibinfo{author}{Tavakoli, A.},
  \bibinfo{author}{Levdik, V.}, \& \bibinfo{author}{Kormushev, P.}
  (\bibinfo{year}{2018}).
\newblock \bibinfo{title}{Time limits in reinforcement learning}.
\newblock In {\it \bibinfo{booktitle}{International Conference on Machine
  Learning}\/} (pp. \bibinfo{pages}{4042--4051}).
\bibitem[{Petit \& Trapp(2019)}]{petit2019enriching}
\bibinfo{author}{Petit, T.}, \& \bibinfo{author}{Trapp, A.~C.}
  (\bibinfo{year}{2019}).
\newblock \bibinfo{title}{Enriching solutions to combinatorial problems via
  solution engineering}.
\newblock {\it \bibinfo{journal}{INFORMS Journal on Computing}\/},  {\it
  \bibinfo{volume}{31}\/}, \bibinfo{pages}{429--444}.
\bibitem[{Popescu et~al.(2021)Popescu, Polat-Erdeniz, Felfernig, Uta, Atas, Le,
  Pilsl, Enzelsberger \& Tran}]{popescu2021overview}
\bibinfo{author}{Popescu, A.}, \bibinfo{author}{Polat-Erdeniz, S.},
  \bibinfo{author}{Felfernig, A.}, \bibinfo{author}{Uta, M.},
  \bibinfo{author}{Atas, M.}, \bibinfo{author}{Le, V.-M.},
  \bibinfo{author}{Pilsl, K.}, \bibinfo{author}{Enzelsberger, M.}, \&
  \bibinfo{author}{Tran, T. N.~T.} (\bibinfo{year}{2021}).
\newblock \bibinfo{title}{An overview of machine learning techniques in
  constraint solving}.
\newblock {\it \bibinfo{journal}{Journal of Intelligent Information
  Systems}\/},  (pp. \bibinfo{pages}{1--28}).
\bibitem[{Refalo(2004)}]{refalo2004impact}
\bibinfo{author}{Refalo, P.} (\bibinfo{year}{2004}).
\newblock \bibinfo{title}{Impact-based search strategies for constraint
  programming}.
\newblock In {\it \bibinfo{booktitle}{International Conference on Principles
  and Practice of Constraint Programming}\/} (pp. \bibinfo{pages}{557--571}).
\newblock \bibinfo{organization}{Springer}.
\bibitem[{Rossi et~al.(2006)Rossi, Van~Beek \& Walsh}]{rossi2006handbook}
\bibinfo{author}{Rossi, F.}, \bibinfo{author}{Van~Beek, P.}, \&
  \bibinfo{author}{Walsh, T.} (\bibinfo{year}{2006}).
\newblock {\it \bibinfo{title}{Handbook of constraint programming}\/}.
\newblock \bibinfo{publisher}{Elsevier}.
\bibitem[{Salido et~al.(2008)Salido, Garrido \&
  Bart{\'a}k}]{salido2008introduction}
\bibinfo{author}{Salido, M.~A.}, \bibinfo{author}{Garrido, A.}, \&
  \bibinfo{author}{Bart{\'a}k, R.} (\bibinfo{year}{2008}).
\newblock \bibinfo{title}{Introduction: Special issue on constraint
  satisfaction techniques for planning and scheduling problems}.
\newblock {\it \bibinfo{journal}{Engineering Applications of Artificial
  Intelligence}\/},  {\it \bibinfo{volume}{21}\/}, \bibinfo{pages}{679--682}.
\bibitem[{Samulowitz \& Memisevic(2007)}]{samulowitz2007learning}
\bibinfo{author}{Samulowitz, H.}, \& \bibinfo{author}{Memisevic, R.}
  (\bibinfo{year}{2007}).
\newblock \bibinfo{title}{Learning to solve qbf}.
\newblock In {\it \bibinfo{booktitle}{Twenty-Second AAAI Conference on
  Artificial Intelligence}\/} (pp. \bibinfo{pages}{255--260}).
\bibitem[{Selsam et~al.(2019)Selsam, Lamm, B{\"u}nz, Liang, de~Moura \&
  Dill}]{selsam2018learning}
\bibinfo{author}{Selsam, D.}, \bibinfo{author}{Lamm, M.},
  \bibinfo{author}{B{\"u}nz, B.}, \bibinfo{author}{Liang, P.},
  \bibinfo{author}{de~Moura, L.}, \& \bibinfo{author}{Dill, D.~L.}
  (\bibinfo{year}{2019}).
\newblock \bibinfo{title}{Learning a sat solver from single-bit supervision}.
\newblock In {\it \bibinfo{booktitle}{International Conference on Learning
  Representations}\/}.
\bibitem[{Song et~al.(2019)Song, Kang, Zhang, Cao \& Xi}]{song2019sampling}
\bibinfo{author}{Song, W.}, \bibinfo{author}{Kang, D.}, \bibinfo{author}{Zhang,
  J.}, \bibinfo{author}{Cao, Z.}, \& \bibinfo{author}{Xi, H.}
  (\bibinfo{year}{2019}).
\newblock \bibinfo{title}{A sampling approach for proactive project scheduling
  under generalized time-dependent workability uncertainty}.
\newblock {\it \bibinfo{journal}{Journal of Artificial Intelligence
  Research}\/},  {\it \bibinfo{volume}{64}\/}, \bibinfo{pages}{385--427}.
\bibitem[{Stuckey et~al.(2014)Stuckey, Feydy, Schutt, Tack \&
  Fischer}]{stuckey2014minizinc}
\bibinfo{author}{Stuckey, P.~J.}, \bibinfo{author}{Feydy, T.},
  \bibinfo{author}{Schutt, A.}, \bibinfo{author}{Tack, G.}, \&
  \bibinfo{author}{Fischer, J.} (\bibinfo{year}{2014}).
\newblock \bibinfo{title}{The minizinc challenge 2008--2013}.
\newblock {\it \bibinfo{journal}{AI Magazine}\/},  {\it
  \bibinfo{volume}{35}\/}, \bibinfo{pages}{55--60}.
\bibitem[{Wu et~al.(2021)Wu, Song, Cao, Zhang \& Lim}]{wu2021learning}
\bibinfo{author}{Wu, Y.}, \bibinfo{author}{Song, W.}, \bibinfo{author}{Cao,
  Z.}, \bibinfo{author}{Zhang, J.}, \& \bibinfo{author}{Lim, A.}
  (\bibinfo{year}{2021}).
\newblock \bibinfo{title}{Learning improvement heuristics for solving routing
  problems}.
\newblock {\it \bibinfo{journal}{IEEE Transactions on Neural Networks and
  Learning Systems}\/},  (pp. \bibinfo{pages}{1--13}).
  \DOIprefix\doi{10.1109/TNNLS.2021.3068828}.
\bibitem[{Xin et~al.(2021{\natexlab{a}})Xin, Song, Cao \& Zhang}]{xin2021multi}
\bibinfo{author}{Xin, L.}, \bibinfo{author}{Song, W.}, \bibinfo{author}{Cao,
  Z.}, \& \bibinfo{author}{Zhang, J.} (\bibinfo{year}{2021}{\natexlab{a}}).
\newblock \bibinfo{title}{Multi-decoder attention model with embedding glimpse
  for solving vehicle routing problems}.
\newblock In {\it \bibinfo{booktitle}{Proceedings of the Thirty-Fifth AAAI
  Conference on Artificial Intelligence}\/}.
\bibitem[{Xin et~al.(2021{\natexlab{b}})Xin, Song, Cao \& Zhang}]{xin2021step}
\bibinfo{author}{Xin, L.}, \bibinfo{author}{Song, W.}, \bibinfo{author}{Cao,
  Z.}, \& \bibinfo{author}{Zhang, J.} (\bibinfo{year}{2021}{\natexlab{b}}).
\newblock \bibinfo{title}{Step-wise deep learning models for solving routing
  problems}.
\newblock {\it \bibinfo{journal}{IEEE Transactions Industrial Informatics}\/},
  {\it \bibinfo{volume}{17}\/}, \bibinfo{pages}{4861--4871}.
  \DOIprefix\doi{10.1109/TII.2020.3031409}.
\bibitem[{Xu et~al.(2018)Xu, Koenig \& Kumar}]{xu2018towards}
\bibinfo{author}{Xu, H.}, \bibinfo{author}{Koenig, S.}, \&
  \bibinfo{author}{Kumar, T.~S.} (\bibinfo{year}{2018}).
\newblock \bibinfo{title}{Towards effective deep learning for constraint
  satisfaction problems}.
\newblock In {\it \bibinfo{booktitle}{International Conference on Principles
  and Practice of Constraint Programming}\/} (pp. \bibinfo{pages}{588--597}).
\newblock \bibinfo{organization}{Springer}.
\bibitem[{Xu et~al.(2007)Xu, Boussemart, Hemery \& Lecoutre}]{xu2007random}
\bibinfo{author}{Xu, K.}, \bibinfo{author}{Boussemart, F.},
  \bibinfo{author}{Hemery, F.}, \& \bibinfo{author}{Lecoutre, C.}
  (\bibinfo{year}{2007}).
\newblock \bibinfo{title}{Random constraint satisfaction: Easy generation of
  hard (satisfiable) instances}.
\newblock {\it \bibinfo{journal}{Artificial intelligence}\/},  {\it
  \bibinfo{volume}{171}\/}, \bibinfo{pages}{514--534}.
\bibitem[{Xu et~al.(2019)Xu, Hu, Leskovec \& Jegelka}]{xu2018powerful}
\bibinfo{author}{Xu, K.}, \bibinfo{author}{Hu, W.}, \bibinfo{author}{Leskovec,
  J.}, \& \bibinfo{author}{Jegelka, S.} (\bibinfo{year}{2019}).
\newblock \bibinfo{title}{How powerful are graph neural networks?}
\newblock In {\it \bibinfo{booktitle}{International Conference on Learning
  Representations}\/}.
\bibitem[{Xu et~al.(2009)Xu, Stern \& Samulowitz}]{xu2009learning}
\bibinfo{author}{Xu, Y.}, \bibinfo{author}{Stern, D.}, \&
  \bibinfo{author}{Samulowitz, H.} (\bibinfo{year}{2009}).
\newblock \bibinfo{title}{Learning adaptation to solve constraint satisfaction
  problems}.
\newblock {\it \bibinfo{journal}{Proceedings of Learning and Intelligent
  Optimization (LION)}\/}, .
\bibitem[{Zarpellon et~al.(2021)Zarpellon, Jo, Lodi \& Bengio}]{jo2021param}
\bibinfo{author}{Zarpellon, G.}, \bibinfo{author}{Jo, J.},
  \bibinfo{author}{Lodi, A.}, \& \bibinfo{author}{Bengio, Y.}
  (\bibinfo{year}{2021}).
\newblock \bibinfo{title}{Parameterizing branch-and-bound search trees to learn
  branching policies}.
\newblock In {\it \bibinfo{booktitle}{Proceedings of the AAAI Conference on
  Artificial Intelligence}\/} (pp. \bibinfo{pages}{3931--3939}).
\newblock volume~\bibinfo{volume}{35}.
\bibitem[{Zhang et~al.(2020)Zhang, Song, Cao \& Zhang}]{zhang2020learning}
\bibinfo{author}{Zhang, C.}, \bibinfo{author}{Song, W.}, \bibinfo{author}{Cao,
  Z.}, \& \bibinfo{author}{Zhang, J.} (\bibinfo{year}{2020}).
\newblock \bibinfo{title}{Learning to dispatch for job shop scheduling via deep
  reinforcement learning}.
\newblock In {\it \bibinfo{booktitle}{Advances in Neural Information Processing
  Systems}\/}.

\end{thebibliography}

\end{document}